\documentclass[runningheads]{llncs}
\usepackage[utf8]{inputenc}

\usepackage{array}
\usepackage{algorithm}
\usepackage{algpseudocode}
\usepackage{amsfonts}
\usepackage{amsmath}
\usepackage{mathtools}
\usepackage{bbm}
\usepackage{bm}
\usepackage{booktabs}
\usepackage{overpic}
\usepackage{graphicx}
\usepackage{multirow}
\usepackage{hyperref}
\usepackage{grffile}

\newcommand{\A}[1]{A^{(#1)}}
\newcommand{\B}[1]{b^{(#1)}}
\newcommand{\hA}[1]{\hat{A}^{(#1)}}
\newcommand{\hB}[1]{\hat{b}^{(#1)}}
\newcommand{\tA}[1]{\tilde{A}^{(#1)}}
\newcommand{\tB}[1]{\tilde{b}^{(#1)}}

    \makeatletter
    \let\@fnsymbol\@arabic
    \makeatother

\newcommand{\relu}{\text{ReLU}}
\newtheorem{defn}{Definition}
\newtheorem{assumption}{Assumption}

\makeatletter
\renewcommand\subsubsection{\@startsection{subsubsection}{3}{\z@}%
                       {-18\p@ \@plus -4\p@ \@minus -4\p@}%
                       {0.5em \@plus 0.22em \@minus 0.1em}%
                       {\normalfont\normalsize\bfseries\boldmath}}
\makeatother
\title{Cryptanalytic Extraction of \\ Neural Network Models}

\author{Nicholas Carlini\thanks{Google} \qquad Matthew Jagielski\thanks{Northeastern University, part of work done at Google} \qquad Ilya Mironov\thanks{Facebook, part of work done at Google}}

\institute{}

\begin{document}

\maketitle

\thispagestyle{plain}

\begin{abstract}
    We argue that the machine learning problem of \emph{model extraction} is actually a
    cryptanalytic problem in disguise, and should be studied as such.
    Given oracle access to a neural network, we introduce a differential attack that
    can efficiently steal the parameters of the remote model up to floating point precision.
    Our attack relies on the fact that ReLU neural networks are piecewise linear
    functions, and thus queries at the critical points reveal information
    about the model parameters.
    
    We evaluate our attack on multiple neural network models and extract models that
    are $2^{20}$ times more precise and require $100\times$ fewer queries than prior work.
    For example, we extract a $100{,}000$ parameter neural
    network trained on the MNIST digit recognition task
    with $2^{21.5}$ queries in under an hour, such that the extracted model agrees
    with the oracle on all inputs up to a worst-case error of $2^{-25}$,
    or a model with $4{,}000$ parameters in $2^{18.5}$ queries
    with worst-case error of $2^{-40.4}$.
    Code is available at \url{https://github.com/google-research/cryptanalytic-model-extraction}.
\end{abstract}

\section{Introduction}
\label{sec:intro}

The past decade has seen significant advances in machine learning, and deep learning in particular.
Tasks viewed as being completely infeasible at the beginning of the decade became almost completely
solved by the end.
AlphaGo \cite{silver2016mastering} defeated 
professional players at Go, a feat in 2014 seen as being at least ten years away~\cite{WiredGo}.
Accuracy on the ImageNet recognition benchmark improved from $73\%$ in 2010 to
$98.7\%$ in 2019, a $20\times$ reduction in error rate~\cite{xie2019self}.
Neural networks can generate photo-realistic high-resolution
images that humans find indistinguishable from actual photographs \cite{Karras2019stylegan2}.
Neural networks achieve higher accuracy than human doctors in limited
settings, such as early cancer detection~\cite{esteva2017dermatologist}.

These advances have brought neural networks into production systems.
The automatic speech recognition systems on
Google's Assistant, Apple's Siri, and Amazon's Alexa are all
powered by speech recognition neural networks.
Neural Machine Translation \cite{bahdanau2014neural} is now the
technique of choice for production language translation systems \cite{wu2016google}.
Autonomous driving is only feasible because of these improved image recognition neural networks.

High-accuracy neural networks are often held secret for at least two reasons.
First, they are seen as a
competitive advantage
and are treated as trade secrets \cite{wenskay1990intellectual};
for example, none of the earlier systems are open-source.
%
%
Second, is seen as
improving both security and privacy to keep these models secret.
With full white-box access it is easy to mount
evasion attacks and generate \emph{adversarial examples} \cite{szegedy2013intriguing,biggio2013evasion}
against, for instance, abuse- or spam-detection models.
Further, white-box access allows \emph{model inversion} attacks~\cite{Fredrikson}:
it is possible to reconstruct identifiable images of specific people
given a model trained to recognize specific human faces.
Similarly, given a language model trained on text containing
sensitive data (e.g., credit card numbers), a white-box attacker can
pull this sensitive data out of the trained model~\cite{carlini2019secret}.

Fortunately for providers of machine learning models, it is often expensive to reproduce a neural network.
There are three reasons for this cost:
first, most machine learning requires extensive training data that can
be expensive to collect;
second, neural networks typically need \emph{hyper-parameter tuning}
requiring training many models to identify the optimal final
model configuration; and
third, even performing a final training run given the collected
training data and correctly configured model is expensive.

For all of the above reasons, it becomes clear that (a) adversaries are motivated for various reasons
to obtain a copy of existing deployed neural networks, and (b)
preserving the secrecy of models is highly important.
In practice companies ensure the secrecy of these models by either releasing 
only an API allowing query access, or releasing on-device models,
but attempting to tamper-proof and obfuscate the source
to make it difficult to extract out of the device.

Understandably, the above weak forms of protection are often seen as insufficient.
The area of ``secure inference'' improves on this by bringing tools from Secure Function Evaluation (SFE), which allows mutually distrustful cooperating parties to evaluate $f(x)$ where $f$ is held by one party and $x$ by the other.
The various proposals often apply fully homomorphic encryption \cite{gentry2009fully,gilad2016cryptonets},
garbled circuits \cite{yao1986generate,riazi2018chameleon},
or combinations of the two \cite{mishra2020delphi}.
Per the standard SFE guarantee, secure inference
``does not hide information [about the function $f$] that is revealed by the result
of the prediction'' \cite{mishra2020delphi}.
However this line of work often implicitly assumes that total leakage from
the predictions is small, and that recovering the function from its output would be difficult.

In total, it is clear that protecting the secrecy of neural network models is
seen as important both in practice and in the academic research community.
This leads to the question that we study in this paper:

\addtolength\leftmargini{0.4in}
\begin{quote}
    \emph{Is it possible to \emph{extract} an identical copy of a neural network
    given oracle (black-box) access to the target model?}
\end{quote}
While this question is not new
\cite{tramer2016stealing,milli2018model,jagielski2019high,rolnick2019identifying},
we argue that model extraction should be studied as a cryptanalytic problem.
To do this, we focus on model extraction in an idealized environment
where a machine learning model is made available as an oracle $\mathcal{O}$
that can be queried, but with no timing or other side channels.
This setting captures that of obfuscated models made public,
prediction APIs, and secure inference.

%

%
%

\newpage

\subsection{Model Extraction as a Cryptanalytic Problem}

The key insight of this paper is that model extraction is closely related to an extremely
well-studied problem in cryptography:
the cryptanalysis of blockciphers.
Informally, a symmetric-key encryption algorithm is a keyed function
$E_k\colon \mathcal{X} \to \mathcal{Y}$
that maps inputs (plaintexts)
$x \in \mathcal{X}$
to outputs (ciphertexts)
$y \in \mathcal{Y}$.
We expect all practically important ciphers to be resistant, at the very least, to key recovery under the adaptive chosen-plaintext attack, i.e.,  
given some bounded number of (adaptively chosen) plaintext/ciphertext pairs
$\{(x_i, y_i)\}$ an encryption algorithm is designed so that the key $k$ cannot be extracted by a computationally-bounded adversary.

Contrast this to machine learning. 
A neural network model is (informally) a parameterized function
$f_\theta\colon \mathcal{X} \to \mathcal{Y}$
that maps input (e.g., images)
$x \in \mathcal{X}$
to outputs (e.g., labels)
$y \in \mathcal{Y}$.
A \emph{model extraction} attack adaptively queries the
neural network to obtain a set of input/output pairs $\{(x_i,y_i)\}$ that
reveals information about the weights $\theta$.
Neural networks are not constructed by design to be resistant to such attacks.

Thus, viewed appropriately,
performing a model extraction attack---learning the weights
$\theta$ given oracle access to the function $f_\theta$---is 
a similar problem to performing a \emph{chosen-plaintext attack}
on a nontraditional ``encryption'' algorithm. 

Given that it took the field of cryptography decades to design encryption
algorithms secure against chosen-plaintext attacks,
it would be deeply surprising if neural networks, where such
attacks are not even considered in their design, were \emph{not}
vulnerable.
Worse, the \emph{primary} objective of cipher design is robustness against such attacks.
Machine learning models, on the other hand, are primarily designed to
be \emph{accurate} at some underlying task,
making the design of chosen-plaintext secure neural networks an even more challenging problem.

There are three differences separating model extraction from standard cryptanalysis
that make model extraction nontrivial and interesting to study.

First, the attack success criterion differs.
While a cryptographic break can be successful even without learning key bits---for
example by distinguishing the algorithm from a pseudo-random function,
only ``total breaks'' that reveal (some of) the actual model parameters
$\theta$ are interesting for model extraction.

Second, the earlier analogy to keyed ciphers is imperfect.
Neural networks typically take high-dimensional inputs
(e.g., images)
and return low-dimensional outputs (e.g., a single probability).
It is almost more appropriate to make an analogy to cryptanalysis of
keyed many-to-one functions, such as MACs.
However, the security properties of MACs are quite different from those
of machine learning models, for example second preimages are expected rather than shunned in neural networks.

Finally, and the largest difference in practice,
is that machine learning models deal in fixed- or floating-point reals rather than finite field arithmetic.
As such, there are many components to our attack that would be significantly
simplified given infinitely precise floating-point math, but given the
realities of modern machine learning, require far more sophisticated attack techniques.

\subsection{Our Results}
We introduce a differential attack that is effective at performing
\emph{functionally-equivalent} neural network model extraction attacks.
Our attack traces the neural network's evaluation on
pairs of examples that differ in a few entries and uses
this to recover the layers (analogous to the rounds of a block cipher)
of a neural network one by one.
%
%
%
To evaluate the efficacy of our attack, we formalize the definition of \emph{fidelity} introduced in
prior work \cite{jagielski2019high} and quantify the degree to which a model extraction attack has succeeded:
\begin{defn}
  Two models $f$ and $g$ are $(\epsilon,\delta)$-\emph{functionally equivalent on $S$}
  if
  \[\textstyle\mathop{\mathrm{Pr}}_{x \in S} \big[\lvert f(x) - g(x) \rvert \le \epsilon \big] \ge 1-\delta. \]
\label{def:ed-fe}
\vspace{-1em}
\end{defn}
%
%
%
%
Table~\ref{tab:results} reports the results of our differential attack across a wide range of model sizes
and architectures, reporting both $(\varepsilon,\delta)$-functional equivalence
on the set $S=[0,1]^{d_0}$, the input space of the model,
along with a direct measurement of $\max\,\, \lvert \theta - \hat\theta \rvert$,
directly measuring the error between the actual model weights $\theta$ and the extracted weights $\hat\theta$
(as described in Section~\ref{app:delta0}).

\begin{table}
    \centering
    \setlength{\tabcolsep}{5pt}
    \begin{tabular}{lrlrrrrr}
    \toprule
     \multirow{1}{*}{Architecture} &
     \multirow{1}{*}{Parameters} & \multirow{1}{*}{Approach} & 
     \multirow{1}{*}{Queries} &
     \multirow{1}{*}{$(\varepsilon, 10^{-9})$} &
     \multirow{1}{*}{$(\varepsilon, 0)$} &
     \multirow{1}{*}{$\max \,\lvert \theta - \hat\theta \rvert$}\\
    \midrule
        784-32-1 & 25,120 & \cite{jagielski2019high} & $ 2^{18.2} $    & $2^{3.2}$   & $2^{4.5}$ & $2^{-1.7}$ \\
                 && Ours                    & $2^{19.2}$  & $2^{-28.8}$ & $2^{-27.4}$ & $2^{-30.2}$\\
        784-128-1 & 100,480 & \cite{jagielski2019high} & $ 2^{20.2}$    & $ 2^{4.8} $   & $ 2^{5.1} $ & $2^{-1.8}$\\
                 && Ours                    & $2^{21.5}$  & $2^{-26.4}$ & $2^{-24.7}$ & $2^{-29.4}$\\
        \midrule
        10-10-10-1 &210 & \cite{rolnick2019identifying}                  & $ 2^{22}$        & $2^{-10.3}$ & $2^{-3.4}$ & $2^{-12}$\\
                   && Ours                  & $ 2^{16.0} $     & $2^{-42.7}$ & $2^{-37.98}$ & $2^{-36}$\\
        10-20-20-1 & 420& \cite{rolnick2019identifying}                  & $ 2^{25} $       & $\infty^\dagger$ & $\infty^\dagger$ & $\infty^\dagger$\\
                   && Ours                  & $ 2^{17.1}$      & $2^{-44.6}$ & $2^{-38.7}$ & $2^{-37}$\\
        \midrule
        40-20-10-10-1 & 1,110 & Ours & $2^{17.8}$ & $2^{-31.7}$ & $2^{-23.4}$ & $2^{-27.1}$\\
        80-40-20-1 & 4,020 & Ours & $2^{18.5}$ & $2^{-45.5}$ & $2^{-40.4}$ & $2^{-39.7}$ \\
    \bottomrule
    \vspace{.2em}
    \end{tabular}
    \caption{Efficacy of our extraction attack which is orders of magnitude more precise than prior work
    and for deeper neural networks orders of magnitude more query efficient.
    Models denoted $a$-$b$-$c$ are \emph{fully connected} neural networks with input dimension $a$,
    one hidden layer with $b$ \emph{neurons}, and $c$ outputs; for formal definitions see Section~\ref{sec:prelim}.
    Entries denoted with a $\dagger$ were unable to recover the network after ten attempts.
    \vspace{-1em}}
    \label{tab:results}
\end{table}

\noindent
The remainder of this paper is structured as follows.
We introduce the notation, threat model, and attacker goals
and assumptions used in Section~\ref{sec:prelim}.
In Section~\ref{sec:attack} we introduce an idealized attack that 
extracts $(0,0)$-functionally-equivalent neural
networks assuming infinite precision arithmetic.
Section~\ref{sec:practical} develops an instantiation of this attack that
works in practice with finite-precision arithmetic to yield $(\varepsilon,\delta)$-functionally
equivalent attacks.

\subsection{Related Work}

Model extraction attacks are classified into two categories \cite{jagielski2019high}:
\emph{task accuracy} extraction and \emph{fidelity} extraction.
The first paper to study task accuracy extraction \cite{tramer2016stealing}
introduced techniques to steal
\emph{similar} models that approximately solve the same underlying
decision task on the natural data distribution, but do not necessarily match the predictions of 
the oracle precisely.
While further work exists in this space \cite{chandrasekaran2018exploring,krishna2019thieves},
we instead focus on fidelity extraction where the adversary
aims to faithfully reproduce the 
predictions of the oracle model, when it is incorrect with respect to the ground truth.
Again, \cite{tramer2016stealing} studied this problem and
developed
(what we would now call) functionally equivalent extraction
for the case of completely linear models.

This attack was then extended by a theoretical result
defining and giving a method for performing functionally-equivalent 
extraction
for neural networks with one layer, assuming
oracle access to the gradients \cite{milli2018model}.
A concrete implementation of this one layer attack that works in practice,
handling floating point imprecision,
was subsequently developed through applying
finite differences to estimate the gradient \cite{jagielski2019high}.
Parallel work to this also extended on these results, focusing on deeper
networks, but required tens to hundreds of millions of
queries~\cite{rolnick2019identifying}; while the theoretical results extended to deep networks,
the implementation in practice only extracts up to the first two
layers.
Our work builds on all of these four results to develop an
approach that is $10^6$ times more accurate, requiring $10^3$ times
fewer queries, and applies to larger models.

Even without query access, it is possible to steal models
with just a cache side-channel \cite{batina2018csi},
although with less fidelity than our attack that we introduce
which are $2^{20}\times$ more precise.
Other attacks target \emph{hyperparameter} extraction---that is, extracting high-level details about the model:
through what method it was trained, if it contains convolutions,
or related questions \cite{wang2018stealing}.
It is further possible to steal hyperparameters with cache
side channels \cite{hong2020how}.

Recent work has studied the learnability of deep neural networks with random weights in the statistical query (SQ) model~\cite{das2020learnability}, showing that learnability drops off exponentially with the depth of the network.
This line of work does not address the \emph{cryptographic} hardness of extraction
in the non-SQ model---precisely the question addressed in this work in the empirical setting.

While not directly related to our problem, it is worth noting that
we are not the first to treat neural networks as just another type of
mathematical function that can be analyzed without any
specific knowledge of machine learning.
Shamir et al.~\cite{Shamir19} explain the existence of adversarial examples \cite{szegedy2013intriguing,biggio2013evasion}, which capture evasion attacks on machine learning classifiers, by considering an abstract model of neural networks.

In a number of places, our attack draws inspiration from the 
cryptanalysis of keyed block-ciphers, most prominently differential
cryptanalysis \cite{biham1991differential}.
We neither assume nor require familiarity with this field, but
the informed reader may enjoy certain parallels.

\section{Preliminaries}
\label{sec:prelim}

This paper studies an abstraction of neural networks as functions
$f\colon\mathcal{X} \to \mathcal{Y}$.
Our results are independent of any methods for selecting the function
$f$ (e.g., stochastic gradient descent),
and are independent of any utility of the function~$f$.
As such, machine learning knowledge is neither expected nor necessary.

\subsection{Notation and Definitions}

\begin{defn}
A $k$-deep neural network $f_\theta(x)$ is a function parameterized by $\theta$ that takes inputs from
an input space $\mathcal{X}$ and returns values in an output space~$\mathcal{Y}$. The function $f$ is composed as a sequence of functions alternating between
linear layers $f_j$ and a nonlinear function (acting component-wise) $\sigma$:
\[f = f_{k+1} \circ \sigma \circ \dots  \circ \sigma \circ  f_2  \circ \sigma \circ f_1. \]
\end{defn}
We exclusively study neural networks over
$\mathcal{X} = \mathbb{R}^{d_0}$
and $\mathcal{Y} = \mathbb{R}^{d_{k}}$. (Until Section~\ref{sec:practical} we assume floating-point numbers can represent $\mathbb{R}$ exactly.)

\begin{defn}
The $j$th \emph{layer} of the neural network $f_j$ is given by
the affine transformation
$f_j(x) = A^{(j)} x + b^{(j)}.$
The \emph{weights} $A^{(j)} \in \mathbb{R}^{d_j \times d_{j-1}}$ is a $d_{j}\times d_{j-1}$ matrix;
the \emph{biases} $b^{(j)} \in \mathbb{R}^{d_j}$ is a $d_j$-dimensional vector.
\end{defn}
While representing each layer $f_j$ as a full matrix
product is the most general definition of a layer, which is called \emph{fully connected},
often layers have more structure.
For example, it is common to use (discrete) \emph{convolutions} in neural networks that operate on images. 
Convolutional layers take the input as a $n\times m$
matrix and convolve it with a kernel, such as a $3\times 3$ matrix.
Importantly, however, it is always possible to
represent a convolution as a matrix product.

\begin{defn}
The \emph{neurons} $\{\eta_i\}_{i=1}^{N}$ are functions
receiving an input and passing it through the \emph{activation function} $\sigma$.
There are a total of $N = \sum_{j=1}^{k-1} d_j$ neurons.
\end{defn}
In this paper we exclusively study the ReLU \cite{nair2010rectified} activation function,
given by $\sigma(x) = \max(x,0)$.
Our results are a fundamental consequence of the fact that ReLU neural networks
are piecewise linear functions.


\begin{defn}
The \emph{architecture} of a neural network captures the structure
of $f$: (a) the number of layers, (b) the dimensions of each layer $\{d_i\}_{i=0}^{k}$, and (c)
any additional constraints imposed on the weights $\A{i}$ and biases $\B{i}$.
\end{defn}
We use the shorthand $a$-$b$-$c$ neural network to denote the sizes of each dimension;
for example a 10-20-5 neural network has input dimension 10, one layer
with 20 neurons, and output dimension 5.
This description completely characterizes the structure of $f$ for fully
connected networks.
%
%
In practice, there are only a few architectures that represent most of the
deployed deep learning models \cite{zoph2016neural}, and developing new architectures is an
extremely difficult and active area in research \cite{he2016deep,szegedy2017inception,tan2019efficientnet}.

\begin{defn}
The \emph{parameters} $\theta$ of $f_\theta$ are the concrete assignments
to the weights $\A{j}$ and biases $\B{j}$, obtained during the
process of \emph{training} the neural network.
\end{defn}
It is beyond the scope of this paper to describe the training process which
produces the parameters $\theta$:
it suffices to know that the process of training is often computationally
expensive and that training is a nondeterministic process, and so training
the same model multiple times will give
different sets of parameters.

\subsection{Adversarial Goals and Resources}
\label{ssec:assumptions}

There are two parties in a model extraction attack: the oracle $\mathcal{O}$ who
returns~$f_\theta(x)$, and the adversary who generates queries $x$ to
the oracle.
\begin{defn}
A \emph{model parameter extraction attack} receives oracle access to 
a parameterized function
$f_\theta$ (in our case a $k$-deep neural network) 
and the architecture of $f$,
and returns a set of parameters $\hat{\theta}$ 
with the 
goal that $f_\theta(x)$ is as similar as possible to $f_{\hat{\theta}}(x)$.
\end{defn}

Throughout this paper we use the $\hat{\_}$ symbol to indicate an extracted parameter. For example, $\hat\theta$ refers to the extracted weights of a model $\theta$.

There is a spectrum of similarity definitions between the extracted weights and the oracle model that prior work has studied~\cite{tramer2016stealing,jagielski2019high,krishna2019thieves};
we focus on the setting where the 
adversarial advantage is defined by $(\varepsilon,\delta)$-functionally
equivalent extraction as in Definition~\ref{def:ed-fe}.

Analogous to cryptanalysis of symmetric-key primitives,
the degree to which a model extraction attack succeeds is determined by 
(a) the number of chosen inputs to the model,
and (b) the amount of compute required.

\paragraph{Assumptions.}
We make several assumptions of the oracle $\mathcal{O}$ and the attacker's knowledge. 
(We believe many of these assumptions are not fundamental and can be relaxed. Removing these assumptions is left to future work.)
\begin{itemize}
    \setlength{\leftmargin}{-10em}
    \item \textbf{Architecture knowledge.}
    We require knowledge of the architecture of the neural network.
    \item \textbf{Full-domain inputs.}
    We feed arbitrary inputs from $\mathcal{X}$.
    \item \textbf{Complete outputs.}
    We receive outputs directly from the model $f$ without further processing (e.g., by returning
    only the most likely class without a score).
    \item \textbf{Precise computations.}
    $f$ is specified and evaluated using 64-bit floating-point arithmetic.
    \item \textbf{Scalar outputs.}
    Without loss of generality we
    require the output dimensionality is 1, i.e., $\mathcal{Y} = \mathbb{R}$.
    \item \textbf{ReLU Activations.}
    All activation functions ($\sigma$'s) are \relu's.\footnote{This is the only assumption
    \emph{fundamental} to our 
    work. Switching to any activation that is not piecewise linear would prevent our attack.
    However, as mentioned, all
    state-of-the-art models use exclusively (piecewise linear generalizations of) the ReLU activation function \cite{szegedy2017inception,tan2019efficientnet}.}
\end{itemize}

\section{Overview of the Differential Attack}
\label{sec:overview}

\begin{figure}
  \centering
  \begin{overpic}[unit=1mm,scale=.7]{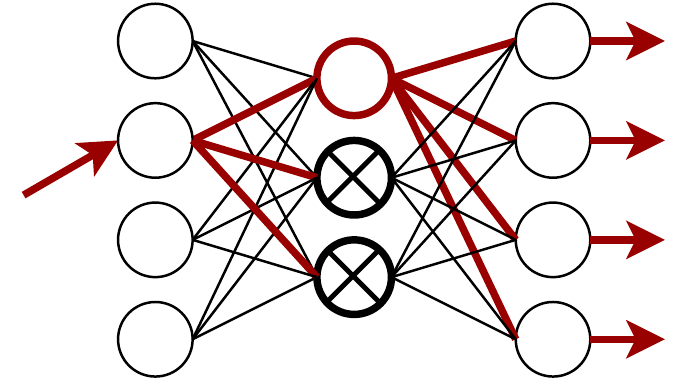}
    \put(-8,19){$\frac{\partial^2 x}{ \partial e_j^2}$}
    \put(100,25){$\frac{\partial^2 f}{\partial x^2} = \A{1}_{ji} \cdot \A{2}_{i}$}
  \end{overpic} 
  \caption{A schematic of our extraction attack on a 1-deep neural network.
    Let $x$ be an input that causes exactly one neuron to have value zero.
    The second differential becomes zero at all other neurons---because they remain
    either fully-inactive or fully-active.
    Therefore the value of this differential is equal to the product of
    the weight going into the neuron at its critical point and the weight
    going out of this neuron.
  }
  \label{fig:schematic}
\end{figure}

Given oracle access to the function $f_\theta$,
we can estimate $\partial f_\theta$ through finite differences along arbitrary
directions.
For simple linear functions defined by $f(x) = a \cdot x + b$,  its directional derivative satisfies
$\frac{\partial f}{\partial e_i} \equiv a_i$, where $e_i$ is the basis vector and $a_i$ is the $i$th entry of the vector $a$, allowing direct recovery of its weights through querying on these well-chosen inputs.

In the case of deep neural networks, we consider second partial directional
derivatives.
ReLU neural networks are piecewise linear functions with
${\partial^2 f \over \partial x^2} \equiv 0$ almost everywhere, except
when the function has some neuron $\eta_j$ at the boundary between the negative and
positive region (i.e., is at its \emph{critical point}).
We show that the value of the partial derivative
$\partial^2 f \over \partial e_i^2$ evaluated 
at a point $x$ so that neuron $\eta_j$ is at such a critical
point actually directly reveals the weight $T(\A{1}_{i,j})$ for some transform
$T$ that is invertible---and therefore the adversary can
learn $\A{1}_{i,j}$.
By repeating this attack along all basis vectors $e_i$ and 
for all neurons $\eta_j$ we can recover the complete matrix $\A{1}$.
Once we have extracted the first layer's weights, we are able to
``peel off'' that layer
and re-mount our attack on the second layer of the neural network,
repeating to the final layer.
There are three core technical difficulties to our attack:

\paragraph{Recovering the neuron signs.}
For each neuron $\eta$, our attack does not exactly recover $\A{l}_i$, the
$i$th row of $\A{l}$, but instead
a scalar multiple $v = \alpha \cdot \A{l}_i$.
While losing a constant $\alpha>0$ keeps the neural network in the same equivalence class,
the sign of $\alpha$ is important and we must distinguish between
the weight vector $\A{l}_i$ and $-\A{l}_i$.
We construct two approaches that solve this problem, but in the general case we
require exponential work (but a linear number of queries).

\paragraph{Controlling inner-layer hidden state.}
On the first layer, we can directly compute the derivative
entry-by-entry, measuring ${\partial^2 f \over \partial e_i^2}$
for each standard basis vector $e_i$ in order to recover $\A{1}_{ij}$.
Deeper in the network, we can not move along 
standard basis vector vectors.
Worse, for each input $x$ on average half of the neurons are in the
negative region and thus their output is identically $0$;
when this happens it is not possible to learn the weight along edges
with value zero.
Thus we are required to develop techniques to elicit behavior from every
neuron, and techniques to cluster together partial recoveries of each row of $\A{l}_i$
to form a complete recovery.

\paragraph{Handling floating-point imprecision.}
Implementing our attack in practice with finite precision neural networks
introduces additional complexity.
In order to estimate the second partial derivative, we require querying on inputs
that differ by only a small amount, reducing the precision of the extracted first
weight matrix to twenty bits, or roughly $10^{-6}$.
This error of $10^{-6}$ is not large to begin with, but this error impacts our ability to recover the next layer, compounding multiplicatively the deeper we go in the network. 
Already in the second layer, the error is magnified to $10^{-4}$, which can completely prevent reconstruction for the third layer:
our predicted view of the hidden state is sufficiently different from the actual hidden state that our attack fails completely.
We resolve this through two means.
First, we introduce numerically stable methods
assuming that all prior layers have been extracted to high precision.
Second, we develop a precision-refinement technique that takes a prefix of the first $j\le k$ layers
of a neural network extracted to $n$ bits of precision
and returns the $j$-deep model extracted to $2n$ bits of precision (up to floating-point tolerance).

\section{Idealized Differential Extraction Attack}
\label{sec:attack}

We now introduce our $(0,0)$-functionally-equivalent model extraction
attack that assumes infinite precision arithmetic and recovers
completely functionally equivalent models.
Recall our attack assumptions (Section~\ref{ssec:assumptions}); using these,
we present our attack beginning with two ``reduced-round'' attacks 
on 0-deep 
(Section~\ref{ssec:nodeep})
and 1-deep (Section~\ref{ssec:onedeep}) neural networks, 
and then proceeding to $k$-deep extraction
for contractive (Section~\ref{ssec:contractive}) and expansive (Section~\ref{sec:expansive})
neural networks.
Section~\ref{sec:practical} refines
this idealized attack to work with finite precision.

\subsection{Zero-Deep Neural Network Extraction}
\label{ssec:nodeep}
Zero-deep neural networks are linear functions
$f(x) \equiv \A{1} \cdot x + \B{1}$.
Querying $d_0$ linearly independent suffices to extract $f$
by solving the resulting linear system.

However let us view this problem differently, to illuminate our attack strategy for deeper networks.
Consider the parallel evaluations $f(x)$ and $f(x+\mathbf{\delta})$, with
\[
f(x+\mathbf{\delta}) - f(x) = \A{1}\cdot (x+\mathbf{\delta}) - \A{1}\cdot x = \A{1}\cdot\mathbf{\delta}.
\]
If $\delta=e_i$, the $i$th standard basis vector of $\mathbb{R}^{d_0}$ (e.g., 
$e_2 = \begin{bmatrix} 0 & 1 & 0 & 0 & \dots & 0\end{bmatrix}$), then
\[
f(x+\delta) - f(x) = \A{1} \cdot \delta = \A{1}_i.
\]
This allows us to directly read off the weights of $\A{1}$.
Put differently, we perform finite differences to estimate the gradient of $f$,  given by $\nabla_x f(x)\equiv\A{1}$.

\subsection{One-Deep Neural Network Extraction}
\label{ssec:onedeep}
Many of the important problems that complicate deep
neural network extraction
begin to arise at 1-deep neural networks.
Because the function is no longer completely linear, we require multiple
phases to recover the network completely.
To do so, we will proceed layer-by-layer, extracting the first layer,
and then use the 0-deep neural network attack
to recover the second layer.

\begin{figure}
    \centering
    
  \begin{overpic}[unit=1mm,scale=.5]{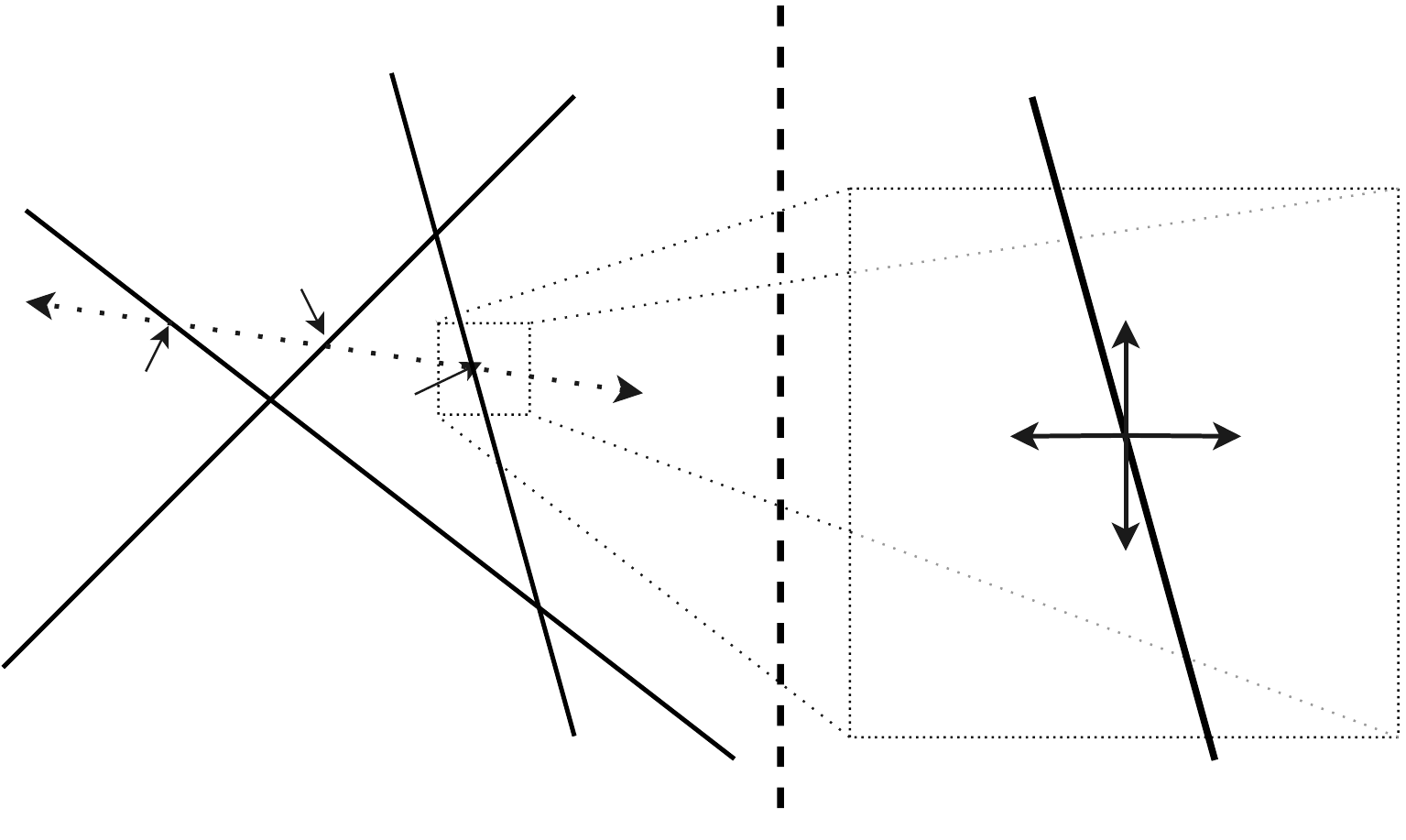}
    \put(25,57){$\eta_0$}
    \put(42,54){$\eta_1$}
    \put(-3,46){$\eta_2$}
    
    \put(-2,37){$\ell$}
    
    \put(70,55){$\eta_0$}
    
    \put(90,28){$e_1$}
    \put(79,39){$e_2$}
    \end{overpic}
    \caption{\textbf{(left)} Geometry of a 1-deep neural network. 
    The three solid lines corresponds to ``critical hyperplanes'' of neurons.
    We identify one \emph{witness} to each neuron with binary
    search on the dotted line $\ell$.
    \textbf{(right)}
    For each discovered critical point, we compute the second partial derivative along
    axis $e_1$ and $e_2$ to compute the angle of the hyperplane.}
    \label{fig:1deep}
\end{figure}
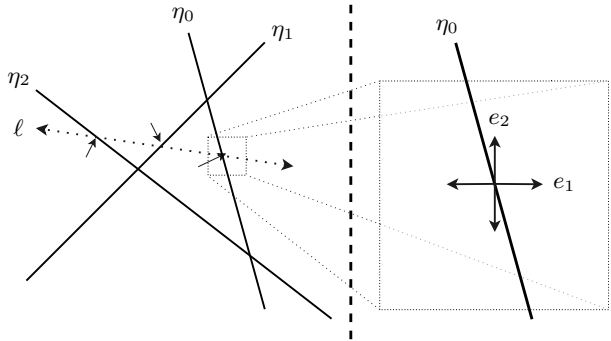

For the remainder of this paper it will be useful to have two distinct mental
models of the problem at hand.
First is the \emph{symbolic} view shown previously in Figure~\ref{fig:schematic}. This
view directly studies the flow of information through the neural networks, represented as an alternating sequence of linear layers and non-linear transformations.
This view helps understanding the algebraic steps of our attack.

The second is the \emph{geometric} view.
Because neural networks operate over the real vector space, they can be 
visualized by plotting two dimensional slices of the landscape \cite{milli2018model}.
Figure~\ref{fig:1deep} (left) contains an example of such a figure.
Each solid black line corresponds to a change in gradient induced in the space
by a neuron changing sign from positive to negative (or vice versa)---ignoring for now the remaining lines.
The problem of neural network extraction corresponds to recovering the
locations and angles of these neuron-induced \emph{hyperplanes}:
in general with input dimension~$d_0$, the planes have dimension $d_0-1$.

\begin{defn}
  The function that computes the first $j$ layers (up to and including~$f_j$
  but not including $\sigma$) of $f$ is denoted
  as $f_{1..j}$. In particular, $f = f_{1..k}$.
\end{defn}

\begin{defn}
The \emph{hidden state} at layer $j$ is the output of the function $f_{1..j}$, before applying the nonlinear transformation $\sigma$.
\end{defn}
Layer $f_j$ is a linear transformation of the $(j-1)$st hidden state after $\sigma$.
\begin{defn}
  $\mathcal{V}(\eta; x)$ denotes the input to neuron $\eta$ (before applying $\sigma$) when evaluated at $x$. 
  $\mathcal{L}(\eta)$ denotes the layer of neuron $\eta$. The first layer starts at 1.
\end{defn}
\begin{defn}
A neuron $\eta$ is at a \emph{critical point} when $\mathcal{V}(\eta; x)=0$.
We refer to this input $x$ as a \emph{witness} to the fact that $\eta$ is at a
critical point, denoted by $x \in \mathcal{W}(\eta)$. 
If $\mathcal{V}(\eta; x) > 0$
then $\eta$ is \emph{active}, and otherwise \emph{inactive}.
\end{defn}
In Figure~\ref{fig:1deep} the locations of these critical points correspond exactly to the
solid black lines drawn through the plane.
Observe that
because we restrict ourselves to ReLU neural networks,
the function $f$ is piecewise
linear and infinitely differentiable almost everywhere.
The gradient $\nabla_x f(x)$ is well defined at all points $x$ except when there exists a neuron
that is at its critical point.

\paragraph{Extracting the rows of $\A{1}$ up to sign.}
Functionally, the attack as presented in this subsection has appeared previously in
the literature \cite{milli2018model,jagielski2019high}.
By framing it differently, our attack will be extensible to deeper
networks.

Assume we were given a witness $x^* \in \mathcal{W}(\eta_j)$ 
that caused neuron $\eta_j$ to be at its critical point 
(i.e., its value is identically zero). 
Because we are using the ReLU activation function, this is the point at
which that neuron is currently ``inactive'' (i.e., is not contributing to the
output of the classifier) but would become ``active'' (i.e., contributing
to the output) if it becomes slightly positive.
Further assume that \emph{only} this neuron $\eta_j$ is at its
critical point, and that for all others neurons $\eta \ne \eta_j$ we
have $|\mathcal{V}(\eta, x_j)| > \delta$ for a constant $\delta > 0$.

Consider two parallel executions of the neural network on pairs of examples.
Begin by defining $e_i$ as the standard basis vectors of $\mathcal{X} = \mathbb{R}^N$.
By querying on the two pairs of inputs $(x^*, x^* + \epsilon e_i)$ 
and $(x^*, x^*- \epsilon e_i)$ we can estimate
\[\alpha^i_+ = {\partial f(x) \over \partial e_i} \bigg \rvert_{x=x^*+\epsilon e_1}
\quad\textrm{and}\quad
\alpha^i_- = {\partial f(x) \over \partial e_i} \bigg \rvert_{x=x^*-\epsilon e_1}
\]
through finite differences.

%

Consider the quantity $\lvert \alpha_+ - \alpha_- \rvert$.
Because $x^*$ induces a critical point of $\eta_j$, 
exactly one of $\{\alpha_+, \alpha_-\}$
will have the neuron $\eta_j$ in its active regime
and the other will have $\eta_j$ in its inactive regime.
If no two columns of $\A{1}$ are collinear, then
as long as $\epsilon < {\delta \over \sum_{i,j} \lvert\A{1}_{i,j}\rvert}$, we are guaranteed that
all other neurons in the neural network will remain in the same state as before---either active
or inactive.
Therefore, if we compute the difference
$\lvert \alpha^i_+ - \alpha^i_- \rvert$,
the gradient information flowing into and out of all other neurons will cancel
and we will be left with just the gradient information flowing along the edge
from the input coordinate $i$ to neuron $\eta_j$ to the output.
Concretely, we can write the 1-deep neural network as
\[f(x) = \A{2} \text{ReLU}(\A{1} x + \B{1}) + \B{2}.\]
and so either $\alpha^i_+ - \alpha^i_- = \A{1}_{j,i} \cdot \A{2}$ or $\alpha^i_- - \alpha^i_+ = \A{1}_{j,i} \cdot \A{2}$.
However, if we repeat the above procedure on a new basis vector
$e_k$ then either
$\alpha^k_+ - \alpha^k_- =  \A{1}_{j,k} \cdot \A{2}$ or $\alpha^k_- - \alpha^k_+ =  \A{1}_{j,k} \cdot \A{2}$ will hold.
Crucially, whichever of the two relations that holds for along coordinate $i$ will be the
same relation that holds on coordinate~$k$.
Therefore we can divide out $\A{2}$ to obtain the ratio of pairs of weights
\[\tfrac{\alpha^k_+ - \alpha^k_-}{\alpha^i_+ - \alpha^i_-}  =  \tfrac{\A{1}_{j,k}}{\A{1}_{j,i}}.\]
This allows us to compute every row of $\A{1}$ up to a single scalar $c_j$.
Further, we can compute $\B{1}_j = -\hA{1}_j \cdot x^*$ (again, up to a scaling factor) because
we know that $x^*$ induces a critical point on neuron $\eta_j$ and so its
value is zero.

%

Observe that the magnitude of $c_j$ is unimportant.
We can always push a constant $c>0$ through to the weight matrix $\A{2}$ and have
a functionally equivalent result.
However, the \emph{sign} of $c_j$ does matter.

\paragraph{Extracting row signs.}
Consider a single witness~$x_i$ for an arbitrary neuron $\eta_i$.
Let $h = f_1(x)$, so that at least one element of $h$ is identically zero.
If we assume that $\A{1}$ is contractive (Section~\ref{sec:expansive} studies non-contractive networks)
then we can find a preimage $x$ to any vector $h$.
In particular, let $e_i$ be the unit vector in the space $\mathbb{R}^{d_1}$.
Then we can compute a preimage $x_+$ so that $\hat{f}_1(x_+) = h + e_i$, and
a preimage $x_-$ so that $\hat{f}_1(x_-) = h - e_i$.

Because $x_i$ is a witness to neuron $\eta_i$ being at its critical point,
we will have that either $f(x_+) = f(x_i)$ or $f(x_-) = f(x_i)$.
Exactly one of these equalities is true because $\sigma(h - e_i) = \sigma(h)$,
but $\sigma(h + e_i) \ne \sigma(h)$ when $h_i = 0$.
Therefore if the second equality holds true, then we know that our extracted
guess of the $i$th row has the correct sign.
However, if the first equality holds true, then our extracted guess
of the $i$th row has the incorrect sign, and so we invert it (along with
the bias $\B{1}_i$).
We repeat this procedure with a critical point for every neuron $\eta_i$ to
completely recover the signs for the full first layer.

\paragraph{Finding witnesses to critical points.}
It only remains to show how to find witnesses $x^* \in \mathcal{W}(\eta)$ for
each neuron $\eta$ on the first layer.
We choose a random line in input space (the dashed line in Figure~\ref{fig:1deep}, left), 
and search along it for nonlinearities in the partial derivative.
Any nonlinearity must have resulted from a ReLU changing signs, 
and locating the specific location where the ReLU changes signs will give us a critical point.
We do this by binary search.

To begin, we take a random initial point $x_0,v \in \mathbb{R}^{d_0}$ together with a large range  $T$.
We perform a binary search for nonlinearities in $f(x_0+tv)$ for $t\in[-T,T]$.
That is, for a given interval $[t_0, t_1]$, we know a critical point exists in the interval if $\frac{\partial f(x+tv)}{\partial v} \rvert_{t=t_0}\neq \frac{\partial f(x+tv)}{\partial v} \rvert_{t=t_1}$.
If these quantities are equal, we do not search the interval, otherwise we continue with the binary search.
%
%
%

\paragraph{Extracting the second layer.}
Once we have fully recovered the first layer weights, we can ``peel off''
the weight matrix $\A{1}$ and bias $\B{1}$ and we are left with extracting the final linear layer,
which reduces to $0$-deep extraction.

\subsection{$k$-Deep Contractive Neural Networks}
\label{ssec:contractive}
Extending the above attack to deep neural networks has several complications that prior work was
unable to resolve efficiently; we address them one at a time.

\paragraph{Critical points can occur due to ReLUs on different layers.} 
Because 1-deep networks have only one layer, all ReLUs
occur on that layer.
Therefore all critical points found during search will
correspond to a neuron on that layer.
For $k$-deep networks this is not true, and if we want to begin by 
extracting the first layer we will have to remove non-first
layer critical points.
(And, in general, to extract layer $j$, we will have to remove non-layer-$j$ critical points.)

\paragraph{The weight recovery procedure requires complete control of the input.}
In order to be able to directly read off the weights, we query the
network on basis vectors~$e_i$.
Achieving this is not always possible for deep networks, and we
must account for the fact that we may only be able to query on
non-orthogonal directions.

\paragraph{Recovering row signs requires computing the preimage of arbitrary hidden states.}
Our row-sign procedure requires that we be able to invert
$\A{1}$, which in general implies we need to develop a
method to compute a preimage of $f_{1..j}$.

\subsubsection{Extracting layer-1 weights with unknown critical point layers}

Suppose we had a function $\mathcal{C}_0(f) = \{x_i\}_{i=1}^{M}$
that returns at least one
critical point for every neuron in the first layer (implying $M \ge d_1$), 
but never returns critical
points for any deeper layer.
We claim that the exact differential attack from above
still correctly recovers
the first layer of a deep neural network.

We make the following observation.
Let $x^* \not\in \bigcup_{\eta_i} \mathcal{W}(\eta_i)$ be an
input that is a witness to no critical point, i.e., $\lvert\mathcal{V}(\eta_i;x^*)\rvert > \epsilon > 0$.
Define $f_{\text{local}}$ as the function so that
for a sufficiently small region we have that $f_{\text{local}} \equiv f$, that is,
\begin{align*}
f_{\text{local}}(x) & = \big(\A{k+1} \cdots (I^{(2)} (\A{2} (I^{(1)} (\A{1}x + \B{1})) + \B{2})) + \dots\big) + \B{k+1} \\
& = \A{k+1} I^{(k)} \A{k} \cdots I^{(2)}  \A{2} I^{(1)} \A{1} x + \beta \\ 
& = \Gamma x + \beta
\end{align*}
Here, $I^{(j)}$ are 0-1 diagonal matrices with a $0$ on the diagonal
when the neuron is inactive and $1$ on the diagonal when
the neuron is active:
\[I^{(j)}_{n,n} = \begin{cases}
1 & \text{if } \mathcal{V}(\eta_n; x) > 0 \\
0 & \text{otherwise}
\end{cases}
\]
where $\eta_n$ is the $n$th neuron on the first layer.
Importantly, observe that each~$I^{(j)}$ is a constant as long as $x$ is 
sufficiently close to $x^*$.
While $\beta$ is unknown, as long as we make only
gradient queries $\partial f_{\text{local}}$, its value is unimportant.
This observation so far follows from the definition  of piecewise linearity.

Consider now some input that is a witness to exactly one
critical point on neuron $\eta^*$. 
Formally, $x^* \in \mathcal{W}(\eta^*)$, 
but $x^* \not\in \bigcup_{\eta_j \ne \eta^*}\mathcal{W}(\eta_j; x^*)$.
Then
\[f_{\text{local}}(x) = \A{k+1} I^{(k)} \A{k} \cdots I^{(2)}  \A{2} I^{(1)}(x) \A{1} x + \beta(x) \]
where again $I^{(j)}$ are 0-1 matrices, but 
except that now, $I^{(1)}$ (and only $I^{(1)}$) is a function of $x$ returning a 0-1 diagonal
matrix that has one of two values, depending on the value of $\mathcal{V}(\eta^*; x) > 0$. 
Therefore we can no longer collapse the matrix product into
one matrix $\Gamma$ but instead can only obtain
\[ f_{\text{local}}(x) = \Gamma I^{(1)}(x) \A{1}x + \beta(x). \]
But this is exactly the case we have already solved for 1-deep neural network
weight recovery: it is equivalent to the statement
$f_{\text{local}}(x) = \Gamma \sigma(\A{1}x + \B{1}) + \beta_2$,
and so by dividing out $\Gamma$ exactly as before
we can recover the ratios of $\A{1}_{i,j}$.

\paragraph{Finding first-layer critical points.}
%
%
Assume we are given a set of inputs $S = \{x_i\}$ so that each
$x_i$ is a witness to neuron $\eta_{x_i}$,
with $\eta_{x_i}$ unknown.
By the coupon collector's argument (assuming uniformity), for $|S|\gg N\log N$, where $N$ is the total number of neurons, we
will have at least \emph{two} witnesses to every neuron~$\eta$.

Without loss of generality let $x_0, x_1 \in \mathcal{W}(\eta)$
be witnesses to the same neuron~$\eta$ on the first layer, i.e, 
that $\mathcal{V}(\eta; x_0) = \mathcal{V}(\eta; x_1) = 0$.
Then, performing the weight recovery procedure beginning from each of these
witnesses (through finite differences) will yield the correct weight vector $\A{1}_j$
up to a scalar.

Typically elements of
$S$ will \emph{not} be witnesses to
neurons on the first layer.
Without loss of generality let $x_2$ and $x_3$ be witnesses to
any neuron on a deeper layer.
We claim that we will be able to detect these error cases:
the outputs of the extraction algorithm will appear to be random and uncorrelated.
Informally speaking, because 
we are running an attack designed to extract first-layer neurons on
a neuron actually on a later layer, it is exceedingly unlikely that the
attack would, by chance, give consistent results when run on $x_2$ and $x_3$
(or any arbitrary pair of neurons).

Formally, let $h_2 = f_1(x_2)$ and $h_3 = f_1(x_3)$.
With high probability, $\text{sign}(h_2) \ne \text{sign}(h_3)$.
Therefore, when executing the extraction procedure on~$x_2$ we compute over the function $\Gamma_1 I^{(1)}(x_2) \A{1}x + \beta_1$,
whereas extracting on~$x_3$ computes over
$\Gamma_2 I^{(1)}(x_3) \A{1}x + \beta_2$.
Because $\Gamma_1 \ne \Gamma_2$, this will give inconsistent results.

Therefore our first layer weight recovery procedure is as follows.
For all inputs $x_i \in S$ run the weight recovery procedure to recover the
unit-length normal vector to each critical hyperplane.
We should expect to see a large number of vectors only once
(because they were the result of running the extraction of a layer $2$ or greater neuron),
and a small number of vectors that
appear duplicated (because they were the result of successful extraction
on the first layer).
Given the first layer, we can reduce the neural
network from a $k$-deep neural network to a $(k-1)$-deep neural network
and repeat the attack. We must resolve two difficulties, however, discussed in the following two subsections.

\subsubsection{Extracting hidden layer weights with unknown critical points}
\label{ssc:unknownlayer}

When extracting the first layer weight matrix,
we were able to compute $\partial^2 f \over \partial e_1 \partial e_j$ for
each input basis vectors $e_i$, allowing us to ``read off'' the ratios of the
weights on the first layer directly from the partial derivatives.
However, for deeper layers, it is nontrivial to exactly control the
hidden layers and change just one coordinate in order to perform finite
differences.\footnote{For the expansive networks we will discuss in Section~\ref{sec:expansive} it is
actually impossible; therefore this section introduces the most general method.}
Let $j$ denote the current layer we are extracting.
Begin by sampling $d_j+1$ directions
$\delta_i \sim \mathcal{N}(0,\epsilon I_{d_0}) \in \mathcal{X}$ and let
\begin{align*}
\{y_i\} = \left\{{\partial^2 f(x) \over \partial \delta_1 \partial \delta_i} \bigg \rvert_{x=x^*}\right\}_{i=1}^{d_j+1}.
\end{align*}
%
%
From here we can construct a system of equations: let $h_i = \sigma(f_{1..j-1}(x + \delta_i))$
and solve for the vector $w$ such that $h_i \cdot w = y_i$.

As before, we run the weight recovery procedure assuming that each witness
corresponds to a critical point on the correct layer.
Witnesses that correspond to neurons on incorrect layers will give
uncorrelated errors that can be discarded.

\paragraph{Unifying partial solutions.}
The above algorithm overlooks one important problem.
For a given critical point $x^*$,
the hidden vector obtained from $f_{1..j}(x^*)$ is likely to have several (on average, half)
neurons that are negative, and therefore $\sigma(f_{1..j}(x^*))$
and any $\sigma(f_{1..j}(x^*+\delta_i))$ will
have neurons that are identically zero.
This makes it impossible to recover the complete weight vector
from just one application of least squares---it is only possible
to compute the weights for those entries that
are non-zero.
One solution would be to search for a witness~$x^*$
such that component-wise $f_{1..j}(x^*) \ge 0$; however
doing this is is not possible in general, and so we do not
consider this option further.

Instead, we combine together multiple attempts at
extracting the weights through a \emph{unification} procedure.
If $x_1$ and $x_2$ are witnesses to critical points for the
same neuron, and the partial vector $f_{1..j}(x_1)$ has entries $t_1 \subset \{1, \dots, d_j\}$ 
and the partial vector $f_{1..j}(x_2)$ has entries $t_2 \subset \{1, \dots, d_j\}$ defined, then
it is possible to recover the ratios for all entries $t_1 \cup t_2$
by unifying together the two partial solutions as long as
$t_1 \cap t_2$ is non-empty as follows.

Let $r_i$ denote the extracted weight vector on witness $x_1$ with
entries at locations $t_1 \subset \{1, \dots, d_j\}$ (respectively, $r_2$ at $x_2$ with locations at~$t_2$).
Because the two vectors correspond to the solution for 
the same row of the weight matrix~$\A{j}_i$, the vectors
$r_1$ and $r_2$ must be consistent on $t_1 \cap t_2$.
Therefore, we will have that $r_1[t_1\cap t_2] = c \cdot r_2[t_1\cap t_2]$
for a scalar $c \ne 0$.
As long as $t_1 \cap t_2 \ne \emptyset$ we can compute
the appropriate constant $c$ and then recover the weight vector
$r_{1,2}$ with entries at positions $t_1 \cup t_2$.

Observe that this procedure also allows us to \emph{check} whether
or not $x_1$ and~$x_2$ are witnesses to the same
neuron $n$ reaching its critical point.
If $|t_1 \cap t_2| \geq \gamma$, then as long as
there do not exist two rows of $\A{j}$ that have $\gamma+1$
entries that are scalar multiples of each other, there will be a unique solution that merges the two partial solutions together.
If the unification procedure above fails---because there does not
exist a single scalar $c$ so that $c \cdot r_1[t_1 \cap t_2] = r_2[t_1 \cap t_2]$---then $x_1$ and $x_2$
are not witnesses to the same neuron being at a critical point.
%

\subsubsection{Recovering row signs in deep networks}
The 1-layer contractive sign recovery procedure
can still apply to ``sufficiently contractive'' neural network 
where at layer $j$ there exists an $\epsilon > 0$ 
so that for all $h \in \mathbb{R}^{d_j}$ with $\lVert h \rVert < \epsilon$
there exists a preimage $x$ with $f_{1..j}(x) = h$.
If a neural network is sufficiently contractive it is easy to see
that the prior described attack will work (because we have assumed
the necessary success criteria).

In the case of $1$-deep networks, it suffices for $d_1 \le d_0$ and
$\A{1}$ to be onto as described.
In general it is necessary that $d_{k} \le d_{k-1} \le \dots \le d_1 \le d_0$ but it
is not sufficient, even if every layer $\A{i}$ were an onto map.
Because there is a ReLU activation after every hidden layer, it is not
possible to send negative values \emph{into} the second layer $f_j$
when computing the preimage.

Therefore, in order to find a preimage of $h_{i} \in \mathbb{R}^{d_i}$
we must be more careful in how we mount our attack: instead of just
searching for $h_{i-1} \in \mathbb{R}^{d_{i-1}}$ so that $f_{i-1}(h_{i-1}) = h_{i}$ we must additionally require
that component-wise $h_{i-1} \ge 0$.
This ensures that we will be able to recursively compute $h_{i-2} \to h_{i-1}$
and by induction compute $x \in \mathcal{X}$
such that $f_{1..j}(x) = h_j$.

%
It is simple to test if a network is sufficiently contractive without
any queries: try the above method to find a preimage $x$; if this fails, 
abort and attempt the following (more expensive) attack procedure. Otherwise
it is contractive.


\subsection{$k$-Deep Expansive Neural Networks}
\label{sec:expansive}

\begin{figure}
    \centering
    
  \scalebox{.9}{
  \begin{overpic}[unit=1mm,scale=.5]{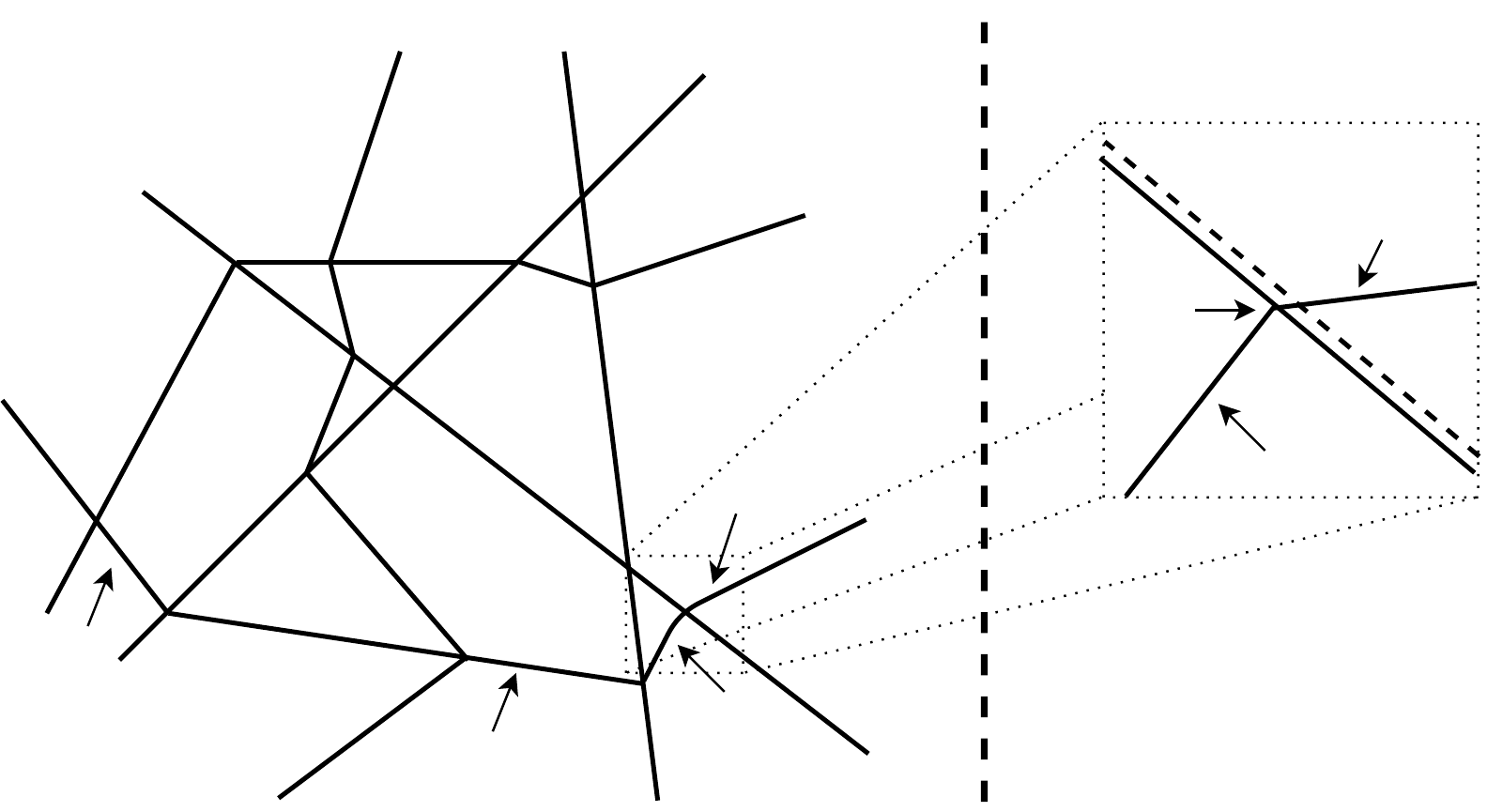}
    \put(35,53){$\eta_0$}
    \put(48,50){$\eta_1$}
    \put(5,43){$\eta_2$}
    
    \put(-4,28){$\eta_3$}
    \put(-1,11){$\eta_4$}
    
    \put(25,53){$\eta_5$}
    
    \put(49,5){$x$}
    \put(49,21){$y$}
    
    \put(77,33){$x'$}
    \put(85,22){$x$}
    \put(93,40){$y$}
    \end{overpic}
    }
    \caption{\textbf{(left)} 
    Geometry of a $k$-deep neural network,
    following~\cite{rolnick2019identifying}.
    Critical hyperplanes induced from neuron $\eta_0, \eta_1, \eta_2$ are on the
    first layer and are linear.
    Critical hyperplanes induced from neurons $\eta_3, \eta_4$ are on the second
    layer and are ``bent'' by neurons on the
    first layer.
    The critical hyperplane induced from neuron $\eta_5$ is a neuron on the third
    layer and is bent by neurons on the prior
    two layers.
    \textbf{(right)}
    Diagram of the hyperplane following procedure. 
    Given an initial witness to a critical point $x$,
    follow the hyperplane to the double-critical point $x'$.
    To find where it goes next, perform binary search along the dashed line and find
    the witness $y$.
    }
    \label{fig:expansive}
\end{figure}
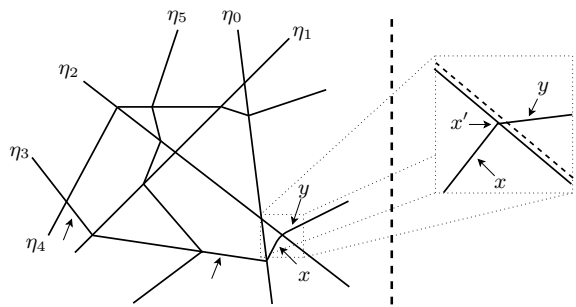

While most small neural networks are contractive, in practice almost
all interesting neural networks are expansive: the number
of neurons on some intermediate layer is larger than the number of
inputs to that layer.
Almost all of the prior methods still apply in this setting, with one
exception:
the column sign recovery procedure.
%
%
Thus, we are required to develop a new strategy.


\paragraph{Recovering signs of the last layer.}
Observe that sign information is not lost for the
final layer: because there is no ReLU activation and we can directly solve
for the weights with least squares, we do not lose sign
information.

\paragraph{Recovering signs on the second-to-last layer.}
Suppose we had extracted completely the function $\hat{f}_{1..k-1}$ (the third
to last layer), and further
had extracted the weights
$\hA{k}$ and biases $\hB{k}$ up to sign of the rows.
There are three unknown quantities remaining: a sign vector
$s \in \{-1,1\}^{d_{k}}$, $\hA{k+1}$ and~$\hB{k+1}$.
Suppose we were given $S \subset \mathcal{X}$ so that $\lvert S \rvert > d_{k}$.
Then it would be possible to solve for all three unknown simultaneously
through brute force.

\begin{defn}
Let
$v \odot M = M'$ denote multiplying rows of matrix $M \in \mathbb{R}^{a \times b}$
by the corresponding coordinate from $v \in \mathbb{R}^a$.
Thus, $M'_{ij} = M_{ij} \cdot v_i$.
\end{defn}

Let $h_i = \sigma(f_{1..k-1}(x_i))$.
Enumerate all $2^{d_{k}}$ assignments of $s$ and compute
$g_i = \sigma( (s \odot \hA{k}) h_i + (s \odot \hB{k})$.
We know that if we guessed the sign vector $s$ correctly, then
there would exist a solution to the system of equations $v \cdot g_i + b = f(x_i)$.
This is the zero-deep extraction problem and solving it
efficiently requires just a single call to
least squares.
This allows us to---through brute forcing the sign bits---completely recover
both the signs of the second-to-last layer as well as the values (and signs)
of the final layer.

Unfortunately, this procedure does not scale to recover the signs of
layer $k-1$ and earlier.
It relies on the existence of an efficient testing procedure (namely, least squares) to solve
the final layer.
If we attempted this brute-force strategy at layer $k-3$ in order to
test if our sign assignment was correct, we would need to run the complete
layer $k-2$ extraction procedure, thus incurring an exponential number of
queries to the oracle.

However, we can use this idea in order to still recover signs even at
earlier layers in the network with only a linear number of queries (but
still exponential work in the width of the hidden layers).

\paragraph{Recovering signs of arbitrary hidden layers.}
Assume that we are given a collection of examples
$\{x_i\} \subset \mathcal{W}(\eta)$ for some neuron $\eta$ that is on the layer after we extracted so far:
$\mathcal{L}(\eta) = j+1$.
Then we would know that there should exist a single unknown vector $v$
and bias $b$
such that $f_j(x_i) \cdot v + b = 0$ for all $x_i$.

This gives us an efficient procedure to test whether or not a given
sign assignment on layer $j$ is correct.
As before, we enumerate all possible sign assignments
and then check if we can recover such a vector $v$. If so, the
assignment is correct; if not, it is wrong.
It only remains left to show how to implement this procedure to
obtain such a collection of inputs $\{x_i\}$.

\subsubsection{The polytope boundary projection algorithm}
\begin{defn}
  The \emph{layer $j$ polytope} containing $x$ is the set of points $\{x + \delta\}$ 
  so that $\text{sign}(\mathcal{V}(\eta; x)) = \text{sign}(\mathcal{V}(\eta; x+\delta))$
  for all $\mathcal{L}(\eta) \le j$.
\end{defn}
Observe that the layer $j$ polytope
around $x$ is an open, convex set, as long as $x$ is not a witness to a critical point.
In Figure~\ref{fig:expansive}, each enclosed region is a layer-$k$ polytope and the triangle formed by $\eta_0, \eta_1$, and $\eta_2$
is a layer-$(k-1)$ polytope.

Given an input $x$ and direction $\Delta$, we can compute the distance
$\alpha$ so that the value $x' = x + \alpha \Delta$ is at the
boundary of the polytope defined by layers $1$ to~$k$.
That is, starting from $x$ traveling along direction $\Delta$ we stop
the first time a neuron on layer $j$ or earlier reaches a critical
point.
Formally, we define
\[ \text{Proj}_{1..j}(x, \Delta) = \textstyle\min_{\alpha \ge 0} \,\, \lbrace \alpha : \exists \eta \,\text{s.t.} \,\, \mathcal{L}(\eta) \le j \wedge \mathcal{V}(\eta; x + \alpha \Delta) = 0 \rbrace \]
We only ever compute $\text{Proj}_{1..j}$
when we have extracted the neural network up to layer $j$.
Thus we perform the computation with respect to the extracted function $\hat{f}$ and
neuron-value function $\hat{\mathcal{V}}$,
and so computing this function requires no queries to the oracle.
In practice we solve for $\alpha$ via binary search.

\subsubsection{Identifying a single next-layer witness}
Given the correctly extracted network $\hat{f}_{1..j-1}$ and the
weights (up to sign) of layer $j-1$, our sign extraction procedure
requires \emph{some} witness
to a critical point on layer $j$.
We begin by performing our standard binary search sweep to find
a collection $S \subset \mathcal{X}$, each of which is a witness
to some neuron on an unknown layer.
It is simple to filter out critical points on layers
$j-1$ or earlier by checking if any of $\hat{\mathcal{V}}(\eta; x)=0$
for $\mathcal{L}(\eta) \le j-1$.
Even though we have not solved for the sign of layer $j$, it is
still possible to compute whether or not they are at a critical
point because critical points of $\hA{j}$ are critical points
of $-\hA{j}$.
This removes any witnesses to critical points on layer $j$ or
lower.

Now we must filter out any critical points on layers strictly later than
$j$.
Let $x^* \in \mathcal{W}(\eta^*)$ denote a potential witness
that is on layer $j$ or later (having already filtered
out critical points on layers $j-1$ or earlier).
Through finite differences, estimate $g = \pm \nabla_x f(x)$ evaluated at $x=x^*$.
Choose any random vector $r$ perpendicular to $g$, and therefore parallel to the critical hyperplane.
Let $\alpha = \text{Proj}_{1..j}(x^*, r)$.
If it turns out that $x^*$ is a witness to a critical point on layer $j$ then
for all $\epsilon < \alpha$ we 
must have that $x^* + \epsilon r \in \mathcal{W}(\eta^*)$.
Importantly, we also have the converse: with high probability for
$\delta > \alpha$ we have that
$x^* + \delta r \not\in \mathcal{W}(\eta^*)$.
However, observe that if $x^*$ is \emph{not} a witness to a neuron
on layer $j$ then one of these two conditions will be false.
We have already ruled out witnesses on \emph{earlier} neuron, 
so if $x^*$ is a witness to a \emph{later} neuron on layer $j'>j$
then it is unlikely that the layer-$j'$ polytope is the same
shape as the layer-$j$ polytope, and therefore we will discover
this fact.
In the case that the two polytopes are actually identical,
we can mount the following attack and if it fails we know
that our initial input was on the wrong layer.

\subsubsection{Recovering multiple witnesses for the same neuron}
\label{sssec:following}
The above procedure yields a single witness $x^* \in \mathcal{W}(\eta^*)$
so that $\mathcal{L}(\eta^*) = j+1$.
We expand this to a collection of witnesses
$W$ where all $x \in W$ have $x \in \mathcal{W}(\eta^*)$, requiring the set to be \emph{diverse}:
%
%
\vspace{-.1em}
\begin{defn}
  A collection of inputs $S$ is \emph{fully diverse} at layer $j$ if
  for all $\eta$ with $\mathcal{L}(\eta) = j$ and for $s \in \{-1,1\}$
  there exists $x \in S$ such that $s \cdot \mathcal{V}(\eta; x) \ge 0$.
\end{defn}
Informally, being diverse at layer $j$ means that if we consider the 
projection onto the space of layer $j$ (by computing $f_{1..j}(x)$ for $x \in S$),
for every neuron $\eta$ there will be at least one input $x_+ \in S$ that
the neuron is positive, and at least one input $x_- \in S$ so that
the neuron is negative.

Our procedure is as follows.
Let $n$ be normal to the hyperplane $x^*$ is on.
Choose some $r$ with $r \cdot n = 0$ and let $\alpha = \text{Proj}_{1..j}(x^*, r)$
to define $x' = x^* + \alpha r$ as a point on the layer-$j$ polytope boundary.
In particular, this implies that we still have that
$x' \in \mathcal{W}(\eta^*)$ (because $r$ is perpendicular to $n$)
but also
$x' \in \mathcal{W}(\eta_u)$
for some neuron $\mathcal{L}(\eta_u) < j$ (by construction of $\alpha$).
Call this input $x'$ the double-critical point (because it is a witness
to two critical points simultaneously).

From this point $x'$, we would like to obtain a new point $y$ 
so that we still have $y \in \mathcal{W}(\eta^*)$, but that also $y$ is on the
other side of the neuron $\eta_u$, i.e.,
$\text{sign}(\mathcal{V}(\eta_u; x^*)) \ne \text{sign}(\mathcal{V}(\eta_u; y))$.
Figure~\ref{fig:expansive} (right) gives a diagram of this process.
In order to follow $x^*$ along its path, we first
need to find it a critical point on the new
hyperplane, having just been bent by the neuron $\eta_u$.
We achieve this by performing a critical-point search
starting $\epsilon$-far away from, and parallel to,
the hyperplane from neuron $\eta_u$ (the dashed line in Figure~\ref{fig:expansive}).
This returns a point $y$ from where we can continue the hyperplane following procedure.

The geometric view hurts us here: because the diagram is a two-dimensional
projection, it appears that from the critical point~$y$
there are only two directions we can travel in: 
\emph{away} from~$x'$ or \emph{towards}~$x'$.
Traveling away is preferable---traveling towards
$x'$ will not help us construct a fully diverse set of
inputs.

However, a $d_0$-dimensional input space has in general $(d_0-1)$ dimensions that
remain on the neuron $\eta^*$.
We defer to Section~\ref{sec:nextdirection} an efficient method for selecting the
continuation direction. For now, observe that 
choosing a random direction will eventually succeed at constructing
a fully-diverse set, but is extremely inefficient:
there exist better strategies than choosing the next
direction.

\subsubsection{Brute force recovery}

Given this collection $S$, we can now---through brute force work---recover
the correct sign assignment as follows.
As described above, compute a fully diverse set of inputs $\{x_i\}$ and 
define $h_i = f_{1..j}(x_i)$. 
Then, for all possible $2^{d_{j}}$ assignments of signs $s \in \{-1,1\}^{d_j}$, compute the guessed
weight matrix $\hA{j}_s = s \odot \hA{j}$.

If we guess the correct vector $s$, then we will be able to compute
$\hat{h_i} = \sigma(\hA{j}_v h_i + \hB{j}_v) = \sigma(\hA{j}_v f_{1..j-1}(x_i) + \hB{j}_v)$ for each $x_i \in S$.
Finally, we know that there will exist a vector $w \ne \mathbb{\vec{0}}$
and bias $\hat{b}$ such that for all $h_i$ we have
$\hat{h_i} w + b = 0$.
As before, if our guess of $s$ is wrong, then with overwhelming probability
there will not exist a valid linear transformation $w,b$.
Thus we can recover sign with a linear number of queries and exponential work.



\section{Instantiating the Differential Attack in Practice}
\label{sec:practical}

The above idealized attack would efficiently extract neural network models
but suffers from two problems.
First, many of the algorithms are not numerically stable and introduce
small errors in the extracted weights.
Because errors
in layer $i$ compound and cause further errors at layers $j>i$, it is
necessary to keep errors to a minimum.
Second, the attack requires more chosen-inputs than is necessary;
we develop new algorithms that require fewer queries or re-use
previously-queried samples.

\paragraph{Reading this section.}
Each sub-section that follows is independent from the surrounding
sub-sections and modifies algorithms introduced in Section~\ref{sec:attack}.
For brevity, we assume complete knowledge of the original algorithm 
and share the same notation.
Readers may find it helpful to review the original algorithm before
proceeding to each subsection.

\subsection{Improving Precision of Extracted Layers}

Given a precisely extracted neural network up to layer $j$
so that $\hat{f}_{1..j-1}$ is functionally equivalent to $f_{1..j-1}$,
but so that weights $\hA{j}$ and biases $\hB{j}$ are imprecisely 
extracted due to imprecision in the extraction attack,
we will now show how to extend this to a \emph{refined}
model $\tilde{f}_{1..j}$ that is functionally equivalent to $f_{1..j}$.
In an idealized environment with infinite precision floating
point arithmetic this step is completely unnecessary;
however empirically this step brings the relative error in the
extracted layer's weights from $2^{-15}$ to $2^{-35}$ or better.

To begin, select a neuron $\eta$ with $\mathcal{L}(\eta) = j$. 
By querying the already-extracted model $\hat{f}_{1..j}$,
analytically compute witnesses
$\{x_i\}_{i=1}^{d_j}$ so that each $x_i \in \hat{\mathcal{W}}(\eta)$.
This requires no queries to the model as we have already extracted
this partial model.

If the $\hA{j}$ and $\hB{j}$ were exactly correct
then $\mathcal{W}(\eta; \cdot) \equiv \hat{\mathcal{W}}(\eta; \cdot)$ and so each computed 
critical point $x_i$ would be exactly a critical point of 
the true model $f$ and so $\mathcal{V}(\eta; x_i) \equiv 0$.
However, if there is any imprecision in the computation, then
in general we will have that $0 < \lvert \mathcal{V}(\eta; x_i) \rvert < \epsilon$
for some small $\epsilon>0$.

Fortunately, given this $x_i$ it is easy to compute $x'_i$ so that
$\mathcal{V}(\eta; x'_i)=0$.
To do this, we sample a random $\Delta \in \mathcal{R}^{d_0}$ and
apply our binary search procedure on the range
$[x_i + \Delta, x_i - \Delta]$.
Here we should select $\Delta$ so that 
$\lVert \Delta \rVert$ is sufficiently small that the only critical
points it crosses is the one induced by neuron $\eta$, but sufficiently large
that it does reliably find the true critical point of $\eta$.

Repeating this procedure for each witness $x_i$ gives a set
of witnesses $\{x'_i\}_{i=1}^{d_j}$ to the same neuron $\eta$.
We compute $h_i = \hat{f}_{1..j-1}(x'_i)$ as the hidden
vector that layer $j$ will receive as input.
By assumption $h_i$ is precise already and so $\hat{f}_{1..j-1} \approx f_{1..j-1}$.
Because $x'_i$ is a witness to neuron $\eta$ having
value zero, we know that that $\A{j}_n \cdot h_i = 0$ where
$n$ corresponds to the row of neuron $\eta$ in $\A{j}$.

Ideally we would solve this resulting system with least squares.
However, in practice, occasionally the conversion from $x \to x'$
fails because $x'$ is no longer a witness to the same neuron $\eta'$.
This happens when there is some other neuron (i.e., $\eta'$) that
is closer to $x$ than the true neuron $\eta$.
Because least squares is not robust to outliers this procedure
can fail to improve the solution.

We take two steps to ensure this does not happen.
First, observe that if $\Delta$ is smaller, the likelihood of capturing
incorrect neurons $\eta'$ decreases faster than the likelihood of capturing
the correct neuron $\eta$. 
Thus, we set $\Delta$ to be small enough that roughly half of the attempts
at finding a witness $x'$ fails.
Second, we apply a (more) robust method of determining the vector
that satisfies the solution of equations \cite{jagielski2018manipulating}.
However, even these two techniques taken together occasionally fail to find
valid solutions to improve the quality.
When this happens, we reject this proposed improvement and keep the
original value.

Our attack could be improved with a solution to the following robust statistics problem:
  Given a (known) set $S \subset \mathbb{R}^N$ such that for some
  (unknown) weight vector $w$ we have
  $\mathop{\text{Pr}}_{x \in S}[\lvert w \cdot x + 1 \rvert \le \epsilon] > \delta$ 
  for sufficiently small $\epsilon$, sufficiently large $\delta > 0.5$, and $\delta|S| > N$,
  efficiently recover the vector $w$ to high precision.

\subsection{Efficient Finite Differences}
Most of the methods in this paper are built on computing
second partial derivatives of the neural network $f$, and therefore
developing a robust method for estimating the gradient is
necessary.
Throughout Section~\ref{sec:attack} we compute the partial derivative
of $f$ along direction $\alpha$ evaluated at $x$ with step size $\varepsilon$ as
%

\[ {\partial{}_{\varepsilon} \over {\partial{}_{\varepsilon} \alpha}} f(x)
\overset{\mathrm{def}}{=}
{{f(x + \varepsilon \cdot \alpha) - f(x)} \over {\varepsilon}}.
\]
%
%
%

To compute the second partial derivative earlier, we computed $\alpha^i_+$ and $\alpha^i_-$
by first taking a step towards $x^*+\epsilon_0 e_1$
for a different step size $\epsilon_0$
and then computed the first partial derivative at this location.
However, with floating point imprecision it is not desirable to have two
step sizes ($\epsilon_0$ controlling the distance away from $x^*$ to step, and
$\epsilon$ controlling the step size when computing the partial derivative).
Worse, we must have that $\epsilon \ll \epsilon_0$
because if ${\partial f \over \partial e_1} \epsilon_0 > {\partial f \over \partial e_i} \epsilon$
then when computing the partial derivative along $e_i$ we may cross
the hyperplane and estimate the first partial derivative incorrectly.
Therefore, instead we compute
\[\alpha^i_+ = {\partial f(x) \over \partial e_i} \bigg \rvert_{x=x^*+\epsilon e_i}
\quad\textrm{and}\quad
\alpha^i_- = {\partial f(x) \over \partial \text{ -}e_i} \bigg \rvert_{x=x^*-\epsilon e_i}
\]
where we both step along $e_i$ and also take the partial derivative along the same $e_i$
(and similarly for -$e_i$).
This removes the requirement for an additional hyperparameter and allows
the step size $\epsilon$ to be orders of magnitude larger, but introduces
a new error: we now lose the relative signs of the entries in the row when performing extraction
and can only recover 
$\left\rvert \A{1}_{i,j}/\A{1}_{i,k} \right\rvert$.

\paragraph{Extracting column signs.}
We next recover the value $\text{sign}(\A{1}_{i,j}) \cdot \text{sign}(\A{1}_{i,k})$.
%
%
Fortunately, the same differencing process allows us to learn this information, using the following observation: if $\A{1}_{i,j}$ and $\A{1}_{i,k}$ have the same sign, then moving in the $e_j+e_k$ direction will cause their contributions to add. If they have different signs, their contributions will cancel each other.
That is, if 
\[\left\lvert {\alpha^{j+k}_+ - \alpha^{j+k}_-}\right\rvert = \left\lvert {\alpha^{j}_+ - \alpha^{j}_-}\right\rvert + \left\lvert {\alpha^{k}_+ - \alpha^{k}_-}\right\rvert,\]
we have that 
\[\left\lvert (\A{1}_{i,j}+\A{1}_{i,k})\cdot \A{2} \right\rvert=\left\lvert \A{1}_{i,j}\cdot\A{2}\right\rvert +\left\lvert \A{1}_{i,k}\cdot \A{2} \right\rvert,\]
and therefore that
%
\[\left\lvert \tfrac{\A{1}_{i,j}}{\A{1}_{i,k}} \right\rvert = \tfrac{\A{1}_{i,j}}{\A{1}_{i,k}}.\]

We can repeat this process to test whether each $\A{1}_{i,j}$ has the same sign as (for example) $\A{1}_{i,1}$.
However, we still do not know whether any single $\A{1}_{i,j}$ is positive or negative---we still must
recover the row signs as done previously.

%
%

\subsection{Finding Witnesses to Critical Points}

\begin{figure}
  \centering

  \scalebox{.9}{
  \begin{overpic}[unit=1mm]{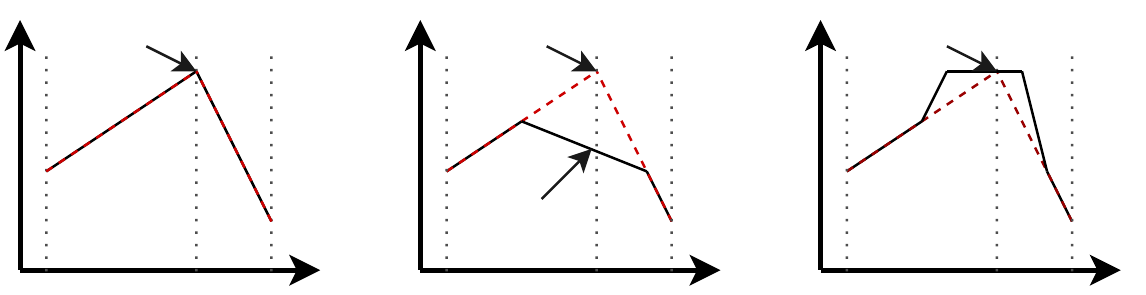}
    \put(3,-1){$x_\alpha$}
    \put(16,-1){$x^*$}
    \put(22,-1){$x_\beta$}

    \put(38,-1){$x_\alpha$}
    \put(51,-1){$x^*$}
    \put(57,-1){$x_\beta$}

    \put(73,-1){$x_\alpha$}
    \put(86,-1){$x^*$}
    \put(92,-1){$x_\beta$}

    \put(4,22){$f(x^*) = \hat{f}(x^*)$}

    \put(74,22){$f(x^*) = \hat{f}(x^*)$}
    
    \put(40,6){$f(x^*)$}
    \put(40,22){$\hat{f}(x^*)$}
  \end{overpic}
  }

  \caption{Efficient and accurate witness discovery. 
  \textbf{(left)} If $x_\alpha$ and $x_\beta$
    differ in only one ReLU (as shown left), we can precisely identify the
    location $x^*$ at which the ReLU reaches its critical point. 
    \textbf{(middle)} If instead
    more than one ReLU differs (as shown right), we can detect that this has
    happened: the predicted of $\hat{f}(\cdot)$ evaluated at $x^*$
    as inferred from intersecting the dotted
    lines does not actually equal the true value of $f(x^*)$.
    \textbf{(right)} This procedure is not \emph{sound} and still may
    potentially incorrectly identify critical points; in practice
    we find these are rare.}
    \label{fig:2linear}
\end{figure}
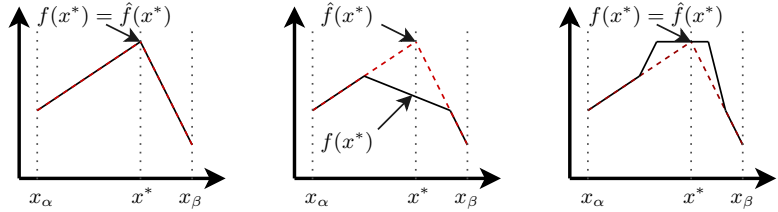

Throughout the paper we require the ability to find witnesses to critical
points.
Section~\ref{ssec:onedeep} uses simple binary search to achieve this which is
(a) imprecise in practice, and (b) query inefficient.
We improve on the witness-finding search procedure developed by \cite{jagielski2019high}.
Again we interpolate between two examples $u,v$ and let $x_\alpha = (1-\alpha) u + \alpha v$.
Previously, we repeatedly performed binary search as long as the partial
derivatives were not equal
$\partial f(x_\alpha) \ne \partial f(x_\beta)$, requiring $p$ queries
to obtain $p$ bits of precision of the value $x^*$.
However, observe that if
$x_\alpha$ and $x_\beta$ differ in the sign of exactly one
neuron $i$, then we can directly compute the location
$x^*$ at which $\mathcal{V}(\eta_i; x^*) = 0$ but so that
for all other $\eta_j$ we have 
\[\text{sign}\big(\mathcal{V}(\eta_j; x_\alpha)\big) = \text{sign}\big(\mathcal{V}(\eta_j; x^*)\big) = \text{sign}\big(\mathcal{V}(\eta_j; x_\beta)\big)\]

This approach is illustrated in Figure~\ref{fig:2linear} and relies on the fact that $f$ is a piecewise linear function with two components.
By measuring, $f(x_\alpha)$ and $\partial f(x_\alpha)$ (resp., $f(x_\beta)$ and $\partial f(x_\beta)$), we find the slope and intercept of both the left and right lines in Figure~\ref{fig:2linear} (left).
This allows us to solve for their expected intersection $(x^*, \hat{f}(x^*))$.
Typically, if there are more than two linear segments, as in the middle of the figure, we will find that the true function value $f(x^*)$ will not agree with the expected function value $\hat{f}(x^*)$ we obtained by computing the intersection; we can then perform binary search again and repeat the procedure.

However, we lose some soundness from this procedure.
As we see in Figure~\ref{fig:2linear} (right), situations may arise where many ReLU units change sign between $x_\alpha$ and $x_\beta$, but $\hat{f}(x^*) = f(x^*)$.
In this case, we would erroneously return $x^*$ as a critical point, and miss all of the other critical points in the range.
Fortunately, this error case is pathological and does not occur in practice.

\subsubsection{Further reducing query complexity of witness discovery}

Suppose that we had already extracted the first $j$ layers of the
neural network and would like to perform the above critical-point
finding algorithm to identify all critical points between $x_\alpha$
and $x_\beta$.
Notice that we do not need to collect any more critical points from the first $j$ layers, but running binary search will recover them nonetheless.
To bypass this, we can analytically compute $S$ as the set of all witnesses to critical points
on the extracted neural network $\hat{f}_{1..j}$ between 
$x_\alpha$ and $x_\beta$.
%
As long as the extracted network $\hat f$ is correct so far, we
are guaranteed that all points in $S$ are also witnesses to critical
points of the true $f$.

Instead of querying on the range $(x_\alpha,x_\beta)$ we perform
the $\lvert S \rvert + 1$ different searches.
Order the elements of $S$ as $\{s_i\}_{i=1}^{|S|}$
so that $s_i < s_j \implies \lvert x_\alpha - s_i \rvert < \lvert x_\alpha - s_j \rvert $.
Abusing notation, let $s_1 = x_\alpha$ and $s_{|S|} = x_\beta$.
Then, perform binary search on each disjoint range $(S_i, S_{i+1})$ for $i=1$ to~$|S|-1$ and return the union.

%
%
%

\subsection{Unification of Witnesses with Noisy Gradients}
Recall that to extract $\hA{l}$ we extract candidates
candidates $\{r_i\}$ and search for pairs $r_i,r_j$ that agree
on multiple coordinates.
This allows us to merge $r_i$ and $r_j$ to recover (eventually) full rows of $\hA{l}$.
With floating point error, the unification algorithm in 
Section~\ref{ssc:unknownlayer} fails for several reasons.

Our core algorithm computes the normal to a hyperplane, returning pairwise ratios 
$\hA{1}_{i,j}/\hA{1}_{i,k}$; throughout Section~\ref{sec:attack}
we set $\hA{1}_{i,1} = 1$ without loss of generality.

Unfortunately in practice there is loss of generality, due to the disparate impact of numerical instability.
Consider the case where $\A{l}_{i,1} < 10^{-\alpha}$ for $\alpha \gg 0$,
but $\A{l}_{i,k} \ge 1$ for all other $k$.
Then there will be substantially more (relative)
floating point imprecision in the weight
$\A{l}_{i,1}$ than in the other weights.
Before normalizing there is no cause for concern since the absolute error
is no larger than for any other.
However, the described algorithm now normalizes every \emph{other} coordinate $\A{l}_{i,k}$
by dividing it by $\A{l}_{i,1}$---polluting the precision of these values.

Therefore we adjust our solution.
At layer $l$, we are given a collection of vectors
$R = \{r_i\}_{i=1}^n$ so that each $r_i$ corresponds to the extraction
of some (unknown) neuron $\eta_{i}$.
First, we need an algorithm to cluster the items into
sets $\{S_j\}_{j=1}^{d_l}$ so that $S_j \subset R$ and so that
every vector in $S_j$ corresponds to one neuron on layer $l$.
We then need to unify each set $S_j$ to obtain the final row of $\hA{l}_j$.

\paragraph{Creating the subsets $S$ with graph clustering.}
Let $r^{(a)}_m \in S_n$ denote the $a$th coordinate of the extracted row $r_m$ from
cluster $n$.
Begin by constructing a graph $G=(V,E)$ where each vector $r_i$ corresponds to
a vertex. 
Let $\delta^{(k)}_{ij} = \lvert r^{(k)}_{i} - r^{(k)}_{j} \rvert$
denote the difference between row $r_i$ and row $r_j$ along axis $k$; then connect
an edge from $r_i$ to $r_j$ when the approximate $\lVert \cdot \rVert_0$ norm is sufficiently large
$\sum_k \mathbbm{1}\left[\delta^{(k)}_{ij} < \epsilon\right] > \log{d_0}$.
We compute the connected components of $G$ and partition each set $S_j$ as one
connected component.
Observe that if $\epsilon=0$ then this procedure is exactly what was described
earlier, pairing vectors whose entries agree perfectly; in practice we find
a value of $\varepsilon=10^{-5}$ suffices.

\paragraph{Unifying each cluster to obtain the row weights.}
We construct the three dimensional $M_{i,a,b} = r^{(i)}_a / r^{(i)}_b$.
Given $M$, the a good guess for the scalar $c_{ab}$ so that
$r^{(i)}_a = r^{(i)}_b \cdot C_{ab}$ along as many coordinates $i$ as 
possible is the assignment $C_{ab} = \text{median}_{i} \,M_{i,a,b}$,
where the estimated error is  $e_{ab} = \text{stdev}_{i} \, M_{i,a,b}$.

If all $r_a$ were complete and had no imprecision then
$C_{ab}$ would have no error and so $C_{ab} = C_{ax} \cdot C_{xb}$.
However because it does have error,
we can iteratively improve the guessed $C$ matrix
by observing that if the error $e_{ax} + e_{xb} < e_{ab}$ then
the guessed assignment $C_{ax} \cdot C_{xb}$ is a better guess
than $C_{ab}$.
Thus we replace $C_{ab} \gets C_{ax} \cdot C_{xb}$ 
and update $e_{ab} \gets e_{ax} + e_{xb}$.
We iterate this process until there is no further improvement.
Then, finally, we choose the optimal dimension
$a = \mathop{\text{arg min}}_a \sum_b e_{ab}$
and return the vector $C_a$.
Observe that this procedure closely follows constructing the union of two
partial entries $r_i$ and $r_j$ except that we perform it along the best
axis possible for each coordinate.

\subsection{Following Neuron Critical Points}

Section~\ref{sssec:following} developed techniques to construct a set of witnesses
to the same neuron being at its critical point.
We now numerically-stabilize this procedure.

As before we begin with an input $x^* \in \mathcal{W}(\eta^*)$
and compute the normal vector $n$ to the critical plane at $x^*$,
and then choose $r$ satisfying $r \cdot n = 0$.
The computation of $n$ will necessarily have some floating point
error, so $r$ will too.

This means when we compute
$\alpha = \text{Proj}_{1..j}(x^*, r)$ and let
$x' = x^* + r \alpha$
the resulting $x'$ will be almost exactly a witness to some neuron 
$\eta_u$ with $\mathcal{L}(\eta_u) < j$,
(because this computation was performed analytically on
a precisely extracted model),
but $x'$ has likely drifted off of the original critical plane 
induced by $\eta^*$.

\begin{figure}[h]
    \centering
    \begin{overpic}[unit=1mm,scale=.7]{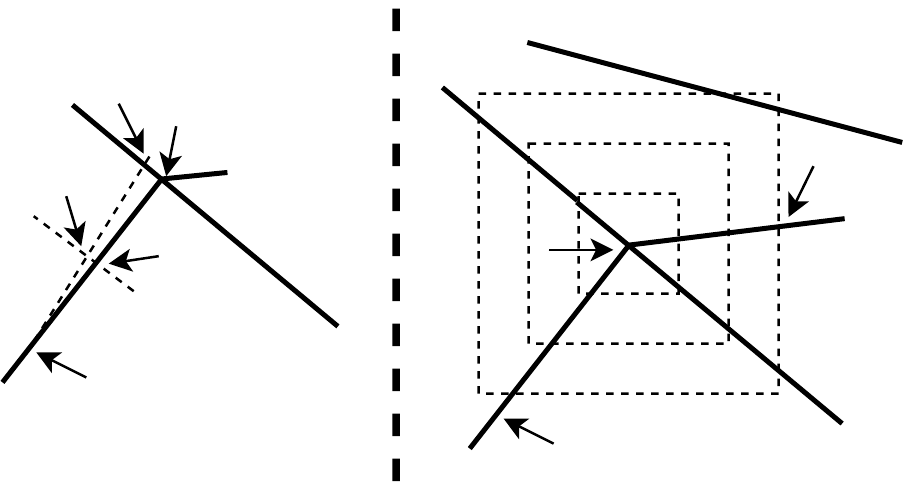}
    \put(2,43){$\eta_u$}
    \put(-5,7){$\eta^*$}
    \put(26,34){$\eta^*$}

    \put(47,0){$\eta^*$}
    
    \put(46,46){$\eta_u$}
    \put(53,50){$\eta_z$}
    \put(94,30){$\eta^*$}

    \put(10,11){$x^*$}
    \put(5,34){$x_1$}
    
    \put(18,25){$x_2$}
    
    \put(11,44){$x'$}
    \put(18,42){$\bar{x}'$}

    \put(63,3){$x^*$}
    \put(53,25){$\bar{x}'$}
    \put(90,37){$y$}
    
    \end{overpic} 
    \caption{Numerically stable critical-point following algorithm.
    \textbf{(left)} From a point $x'$ compute a parallel direction along
    $\eta^*$, step part way to $x_1$ and refine it to $x_2$, and then 
    finish stepping to $x'$.
    \textbf{(right)} From $x'$ grow increasingly large squares until
    there are more than four intersection points; return $y$ as the
    point on $\eta^*$ on the largest square.}
    \label{fig:my_label}
\end{figure}
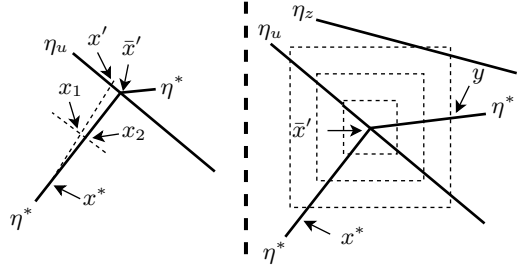

To address this, after computing $\alpha$ we initially take
a smaller step and let $x_1 = x^* + r \sqrt{\alpha}$.
We then refine the location of this point to a point 
$x_2$ by performing binary search on the region $x_1 - \epsilon n$ to
$x_1 + \epsilon n$ for a small step $\epsilon$.
If there was no error in computing $n$ then $x_1 = x_2$ because
both are already witnesses to $\eta^*$.
If not, any error has been corrected.
Given $x^*$ and $x_2$ we now can now compute
$\alpha_2 = \text{Proj}_{1..j}(x^*, x_2-x^*)$
and let $\bar{x}' = x^* + (x_2 - x^*)\alpha_2$ which
will actually be a witness to both neurons simultaneously.

Next we give a stable method to compute $y$
that is a witness to $\eta^*$ and on the other side of $\eta_u$.
The previous procedure required a search parallel to $\eta_u$ and
infinitesimally displaced, but this
is not numerically stable without accurately yet knowing
the normal to the hyperplane given by $\eta_u$.

Instead we perform the following procedure.
Choose two orthogonal vectors of equal length $\beta$, $\gamma$ and
and perform binary search on the line segments
that trace out the perimeter of a square with
coordinates $\bar{x}' \pm \beta \pm \gamma$.

When $\lVert \beta \rVert$ is small, the number of 
critical points crossed will be exactly four: two 
because of $\eta_u$ and two because of $\eta^*$.
As long as the number of critical points remains four, we double
the length of $\beta$ and~$\gamma$.

Eventually we will discover more than four critical points, when
the perimeter of the square intersects another neuron $\eta_z$.
At this point we stop increasing the size of the box and can
compute the continuation direction of $\eta^*$
by discarding the points that fall on $\eta_u$.
We can then choose $y$ as the point on $\eta^*$ that intersected
with the largest square binary search.

\subsubsection{Determining optimal continuation directions}
\label{sec:nextdirection}

The hyperplane following procedure will \emph{eventually} recover a
fully diverse set of inputs $W$ but it may take a large number of
queries to do so.
We can reduce the number of queries by several orders of magnitude
by carefully choosing the continuation direction $r$ instead of
randomly choosing any value so that $r \cdot n = 0$.

Given the initial coordinate $x$ and after computing the normal
$n$ to the hyperplane, we have $d_0-1$ dimensions that we can
choose between to travel next.
Instead of choosing a random $r \cdot n = 0$ we instead choose
$r$ such that we make progress towards obtaining a fully
diverse set $W$.

Define $W_i$ as the set of witnesses that have been found so far.
We say that this set is diverse on neuron $\eta$ if there exists
an $x_+, x_- \in W_i$ such that $\mathcal{V}(\eta; x_+) \ge 0$
and $\mathcal{V}(\eta; x_-) < 0$.
Choose an arbitrary neuron $\eta_t$ such that $W_i$ is not
diverse on~$\eta_t$.
(If there are multiple such options, we should prefer
the neuron that would be \emph{easiest} to reach, but this is secondary.)

Our goal will be to choose a direction $r$ such that
(1) as before, $r \cdot n = 0$, however
(2) $W_i \cup \{x + \alpha r\}$ is closer to being fully
diverse.
Here, ``closer'' means that $d(W) = \min_{x \in W} \lvert \mathcal{V}(\eta_t; x) \rvert$ is smaller.
Because the set is not yet diverse on $\eta_t$, all values are
either positive or negative, and it is our objective to
switch the sign, and therefore become closer to zero.
Therefore our procedure sets
$$r = \textstyle\mathop{\text{arg min}}_{r\colon r \cdot n = 0} d(W_i \cup \{x + \alpha r\})$$
\vspace{-1mm}
performing the minimization through random search over $1{,}000$ directions.

\section{Evaluation}

We implement the described extraction algorithm in
JAX~\cite{jax2018github}, a
Python library that mirrors the NumPy interface 
for performing efficient numerical computation
through just in time compilation.
%

\subsection{Computing $(\varepsilon,10^{-9})$-Functional Equivalence}

Computing $(\varepsilon,10^{-9})$-functional equivalence is simple.
Let $\bar{S} \subset S$ be a finite set consisting of
$\lvert\bar{S}\rvert>10^9$ different inputs drawn $x \in S$.
Sort $\bar{S}$ by $\lvert f(x) - \hat{f}(x)\rvert$ and choose the lowest $\varepsilon$
so that 
  \[\textstyle\mathop{\mathrm{Pr}}_{x \in \bar{S}} \big[\lvert f(x) - g(x) \rvert \le \epsilon \big] \ge 1-\delta. \]
In practice we set $\lvert\bar{S}\rvert = 10^{9}$ and compute the $\max$
so that evaluating the function is possible under an hour per neural network.

\subsection{Computing $(\varepsilon,0)$-Functional Equivalence}
\label{app:delta0}
Directly computing $(\varepsilon,0)$-functional equivalence is infeasible,
and is NP-hard (even to approximate) by reduction to Subset Sum \cite{jagielski2019high}.
We nevertheless propose two methods that efficiently give upper bounds that perform well.

\paragraph{Error bounds propagation}
The most direct method to compute $(\varepsilon,0)$-functional equivalence
of the extracted neural network
$\hat{f}$ is to compare the weights $\A{i}$ to the weights $\hA{i}$ and
analytically derive an upper bound on the error when performing
inference.
Observe that
(1) permuting the order of the neurons in the network does not change
the output, and (2) any row can be multiplied by a positive scalar $c>0$
if the corresponding column in the next layer is divided by~$c$.
Thus, before we can compare $\hA{i}$ to $\A{i}$ we must ``align'' them.
We identify the permutation mapping the rows of $\hA{l}$ to
the rows of $\A{l}$ through a greedy matching algorithm, and then
compute a single scalar per row $s \in \mathbb{R}_+^{d_i}$.
%
%
To ensure that multiplying by a scalar does not change the output of
the network, we multiply the columns of the next layer $\hA{l+1}$ by
$1/s$ (with the inverse taken pairwise).
The process to align the bias vectors $\B{l}$ is identical,
and the process is repeated for each further layer.

This gives an aligned $\tA{i}$ and $\tB{i}$ from which we can
analytically derive upper bounds on the error.
Let $\Delta_i = \tA{i} - \A{i}$, and let
$\delta_i$ be the largest singular value of $\Delta_i$.
If the $\ell_2$-norm of the maximum error going into layer $i$ is
given by $e_i$ then we can bound the maximum error going
out of layer $i$ as
\[e_{i+1} \le \delta_i \cdot e_i + \lVert \tB{i} - \B{i} \rVert_2.\]
By propagating bounds layer-by-layer we can obtain an upper
bound on the maximum error of the output of the model.

This method is able to prove an upper bound on 
$(\epsilon,0)$ functional equivalence for
some networks, when the pairing algorithm succeeds.
However, we find that there are some networks that are $(2^{-45},10^{-9})$
functionally equivalent but where the weight alignment procedure fails.
Therefore, we suspect that
there are more equivalence classes of functions than scalar
multiples of permuted neurons, and so
develop further methods for tightly computing $(\varepsilon,0)$ functional equivalence.

\paragraph{Error overapproximation through MILP}
The above analysis approach is loose.
Our second approach gives exact bounds with an additive error at most $10^{-10}$.

Neural networks are piecewise linear functions, and so can be cast as a mixed integer linear programming (MILP) problem 
\cite{katz2017reluplex}.
We directly express Definition~\ref{def:ed-fe}
as a MILP, following the process of \cite{katz2017reluplex} by
encoding linear layers directly, and encoding ReLU layers by
assigning a binary integer variable to each ReLU.
Due to the exponential nature of the problem, this approach
is limited to small networks.

State-of-the-art
MILP solvers offer a maximum (relative, additive) error tolerance of $10^{-10}$;
for our networks the
SVD upper bound is often $10^{-10}$ or better, so the MILP solver
gives a \emph{worse} bound, despite theoretically being tight.

\section{Results}
We extract a wide range of neural network architectures;
key results are given in Table~\ref{tab:results} (Section~\ref{sec:intro}).
We compute $(\varepsilon, \delta)$-functional equivalence
at $\delta=10^{-9}$ and $\delta=0$
on the domain $S=\{x\colon \lVert x \rVert_2 < d_0 \,\, \wedge \,\, x \in \mathcal{X}\}$, sufficient to explore both
sides of every neuron.
%
%
%

\section{Concluding Remarks}

We introduce a cryptanalytic method for extracting the weights of a neural network
by drawing analogies to cryptanalysis of keyed ciphers.
Our differential attack requires multiple orders of magnitude fewer queries per
parameter than prior work and extracts models that are multiple orders of 
magnitude more accurate than prior work.
In this work, we do not consider defenses---promising approaches include detecting when an attack is occuring, adding noise at some stage of the model's computation, or only returning the label corresponding to the output, any of these easily break our presented attack.

The practicality of this attack has implications for many areas of
machine learning and cryptographic research.
The field of secure inference relies
on the assumption that observing the output of a neural network does not
reveal the weights.
This assumption is false, and therefore the field of secure inference will
need to develop new techniques to protect the secrecy of trained models.

We believe that by casting neural network extraction
as a cryptanalytic problem, even more advanced cryptanalytic techniques
will be able to greatly improve on our results, reducing the computational complexity, reducing the query complexity and reducing the number of assumptions necessary.

\section*{Acknowledgements}

We are grateful to the anonymous reviewers, Florian Tram\`er, Nicolas Papernot, Ananth Raghunathan, and \'Ulfar Erlingsson for helpful feedback.

\bibliographystyle{alpha}
\bibliography{references}

\end{document}